\documentclass[a4paper]{article}

\usepackage{a4wide}
\usepackage{RR}
\usepackage[utf8]{inputenc}
\usepackage[american]{babel}
\usepackage{microtype}
\usepackage{amsmath,amssymb,amsthm}
\usepackage{tikz}
\usepackage{graphicx}
\usepackage{url}
\usepackage{mathrsfs}
\usepackage[ruled,vlined]{algorithm2e}
\usepackage{figlatex}
\usepackage{subfigure}
\graphicspath{{figures/}}

\newcommand\N{\mathbb{N}}

\newcommand\E{\mathbb{E}}
\renewcommand\P{\mathbb{P}}

\newcommand{\A}{\mathcal{A}}    
\renewcommand{\SS}{\mathcal{S}} 

\newcommand{\PP}{\mathcal{P}}
\newcommand{\GG}{\mathcal{G}}
\newcommand{\FF}{\mathcal{F}}

\def\Reals{\mathbb{R}}

\def\Nats{\mathbb{N}}
\newcommand{\calF}{\mathcal F}

\newcommand{\ind}[1]{\mathbf{1}_{#1}}

\newcommand{\bydef}{\stackrel{\rm{def}}{=}}

\newcommand{\eqdef}{\stackrel{\mathrm{def}}{=}} 

\newtheorem{lem}{Lemma}
\newtheorem{theorem}{Theorem}

\newtheorem{coro}[theorem]{Corollary}

\newcommand{\nin}{\not\in} 

\newcommand{\esp}[1]{\E\left(#1\right)}
\newcommand{\espc}[2]{\E\left(\left.#1 \right| #2\right)}

\newcommand{\abs}[1]{\left| #1 \right|}
\newcommand{\norm}[1]{\left\|#1\right\|}

\newcommand{\normsup}[1]{\left\|#1\right\|_\infty}
\newcommand{\floor}[1]{\left\lfloor#1\right\rfloor}

\def\be{\begin{equation}}
\def\ee{\end{equation}}
\def\ben{\[}
\def\een{\]}
\def\bearn{\begin{eqnarray*}}
\def\eearn{\end{eqnarray*}}
\def\bear{\begin{eqnarray}}
\def\eear{\end{eqnarray}}
\newcommand{\lp}{\left(}
\newcommand{\lb}{\left[}
\newcommand{\lc}{\left\{}

\newcommand{\rp}{\right)}
\newcommand{\rb}{\right]}
\newcommand{\rc}{\right\}}

\newcommand{\eps}{\epsilon}
\newcommand{\cro}[1]{\langle#1\rangle}   

\newcommand{\pder}[2]{\frac{\partial#1}{\partial#2}} 

\newcommand{\sref}[1]{Section~\ref{#1}}
\newcommand{\coref}[1]{Corollary~\ref{#1}}
\newcommand{\eref}[1]{Eq.(\ref{#1})}

\newcommand{\lref}[1]{Lemma~\ref{#1}}
\newcommand{\thref}[1]{Theorem~\ref{#1}}

\RRdate{Avril 2010}
\RRversion{3}       
\RRdater{May 2011} 

\RRauthor{Nicolas Gast\and Bruno Gaujal\and Jean-Yves Le Boudec}
\authorhead{N.Gast \& B. Gaujal \& J.-Y. Le Boudec}

\RRtitle{Modèles Champ Moyen et Processus de Décision Markovien: de
  l'optimisation discrète à l'optimisation continue.}
\RRetitle{Mean field for Markov Decision Processes: from Discrete to
  Continuous  Optimization}
\titlehead{Mean field for Markov Decision Processes}

\RRresume{Ce document étudie la convergence de processus de décision markoviens
composés d'un grand nombre d'objets vers des problèmes d'optimisation sur
des équations différentielles. Nous montrons que le gain optimal du
processus de décision converge vers la solution d'une équation continue de
type ``Hamilton-Jacobi-Bellman''.  La preuve utilise \`a la fois  des outils
classiques des modèles champs moyens et diff\'erents nouveaux  couplages
entre les mod\`eles discrets et continus qui permettent de donner  des bornes explicites.
La méthode est ensuite illustrée par trois exemples concernant des
stratégies d'investissement, du contrôle de dynamiques de population et un
problème d'allocation de ressources.}
\RRmotcle{Champ Moyen, Hamilton-Jacobi-Bellman, Contrôle Optimal, Processus 
  de Décision Markovien}

\RRabstract{We study the convergence of Markov decision processes, composed of a large
number of objects, to optimization problems on ordinary differential
equations. We show that the optimal reward of such a Markov decision
process, which satisfies a Bellman equation, converges to the solution of a
continuous Hamilton-Jacobi-Bellman (HJB) equation based on the mean field
approximation of the Markov decision process. We give bounds on the
difference of the rewards and an algorithm for deriving an
approximating solution to the Markov decision process from a solution of
the HJB equations. We illustrate the method on three examples pertaining,
respectively, to investment strategies, population dynamics control and
scheduling in queues.  They are used to illustrate and justify the
construction of the controlled ODE and to show the advantage of  solving
a continuous HJB equation rather than a large discrete Bellman equation.
}
\RRkeyword{Mean Field, Hamilton-Jacobi-Bellman,
  Optimal Control, Markov Decision Process}

\RRprojet{MESCAL}
\RRtheme{\THNum} 
\RRtheme{\THNum} 
\URRhoneAlpes

\newenvironment{IEEEproof}{
  \begin{proof}
  }{
  \end{proof}
}

\begin{document}
\RRNo{7239}

\makeRR

\section{Introduction}

In this paper we study dynamic optimization problems on Markov
decision processes  composed of a large number of interacting objects.

Consider a system of $N$ objects evolving in a common
environment. At each time step, objects change their state
randomly according to some probability kernel $\Gamma^N$. This
kernel depends on the number of objects in each state, as well as on  the decisions of a centralized controller. Our goal is to study the behavior of the controlled system when
$N$ becomes large.

Several papers investigate  the asymptotic behavior of such systems,
but without controllers.  For example, in
\cite{benaim-leboudec,leboudec-mcdonald-mundinger}, the authors
show that under mild conditions, as $N$ grows,  the system converges to a
deterministic limit. The limiting system can be
of two types, depending on the intensity $I(N)$ (the intensity
is the probability than an object changes its state between two
time steps). If $I(N)=O_{N\to\infty}(1)$, the system converges
to a dynamical system in discrete time
\cite{leboudec-mcdonald-mundinger}. If $I(N)$ goes to $0$ as
$N$ grows, the limiting system is a continuous time dynamical
system and can be described by ordinary differential equations
(ODEs).

\subsection*{Contributions}

Here,  we
consider a Markov decision process where at each time step, a
central controller chooses an action from a predefined set that
will modify the dynamics of the system the controller receives a
reward depending on the current state of the system and on the
action. The goal of the controller is to maximize the expected
reward over a finite time horizon.
We show that when $N$ becomes large this
problem converges to an  optimization problem on an ordinary
differential equation.

More precisely, we focus on the case where the Markov decision process
is such that its  empirical occupancy  measure is also Markov; this occurs when the system consists of many interacting objects, the objects can be observed only through their state and the system evolution depends only on the collection of all states. We show that the optimal reward converges to the optimal reward of the mean
field approximation of the system, which is given by the solution of an HJB equation.
Furthermore, the optimal policy of the mean
field approximation
is also asymptotically optimal in $N$, for the original discrete
system. Our method relies on bounding techniques used in
stochastic approximation and learning
\cite{benveniste1990adaptive,benaim1999dsa}. We also introduce
an original coupling method, where, to each sample path of the
Markov decision process, we associate a random trajectory that is
obtained as a solution of the ODE, i.e. the mean field limit,
controlled by random actions.

This convergence result has an algorithmic by-product. Roughly speaking, when confronted with a large Markov decision problem,
we can first solve the HJB equation for the  associated mean field
limit and then build a decision policy for the initial
system that is asymptotically optimal in $N$.

Our results have two main implications.
The first is to justify the construction of controlled ODEs as
good approximations  of large discrete controlled systems.
This construction is given done without rigorous  proofs. In \sref{sec:vaccination} we
illustrate this point with an example of malware infection in computer systems.

The second implication concerns the effective computation of an
optimal control policy.
In the discrete case, this is usually done by using dynamic
programming for the finite horizon case or by computing a fixed point
of the Bellman equation in the discounted case.
Both approaches suffer from the curse of dimensionality, which makes
them impractical when the state space is too large.
In our context, the size of the state space is exponential in $N$,
making the problem even more acute.
In practice, modern supercomputers only allow us to tackle such optimal
control problems  when
$N$ is no larger than a few tens \cite{Papadimitriou}.

The mean field approach offers an alternative to  brute force
computations. By letting $N$ go to infinity, the discrete problem is
replaced  by a limit
Hamilton-Jacobi-Bellman equation that is deterministic
where the dimensionality  of the original system has been
hidden in the occupancy measure. Solving the HJB equation
numerically is sometimes rather easy, as in the examples in
Sections \ref{sec:harvest}  and \ref{sec:vaccination}. It
provides a deterministic optimal policy whose reward with a
finite (but large) number of objects is remarkably close to the
optimal reward.

\subsection*{Related Work}
Several  papers in the literature are concerned with the problem of
mixing the limiting behavior of  a large number of objects with
 optimization.

In \cite{chen2009approximate}, the value function
of the Markov decision process is approximated by a linearly
parametrized class of functions and a fluid approximation of
the MDP is used. It is shown that a solution of the HJB
equation is a value function for a modification of the
original MDP problem. In
\cite{tsitsiklis1997analysis,de2003linear}, the curse of
dimensionality of dynamic programming is circumvented by
approximating the value function by linear regression.
Here, we use instead a mean field limit approximation
 and prove  asymptotic optimality in $N$  of limit policy.

In \cite{gast-gaujal}, the authors also consider Markov decision processes with a growing number of objects, but
when the intensity is $O(1)$. In their case, the optimization
problem of the system of size $N$ converges to a deterministic
optimization problem in discrete time.
In this paper however, we focus on the $o(1)$ case, which is
substantially different from the discrete time case because the limiting system does not evolve in discrete time anymore.

Actually, most of the papers  dealing with mean field limits of optimization problems over large systems are set in a game theory framework, leading to the concept of   {\it mean field games} introduced in \cite{Lions}. The  objects composing the system are seen as  $N$ players  of a game with distributed information, cost and control; their actions lead to a Nash equilibrium. To the best of our knowledge, the classic case with global information and  centralized control has not yet been considered.
Our work focuses precisely on classic Markov decision problems, where a central controller (our objects are passive), aims at minimizing a global cost function.

For example, a  series of papers by M. Huang, P.E. Caines and  P.  Malham\'e such as \cite{caines1,caines4,caines2,caines3}  investigate  the behavior of systems made of a large number of objects under {\it distributed } control. They mostly investigate Linear-Quadratic- Gaussian (LQG) dynamics and use the fact that, here, the solution can be given in closed form as a Riccati equation  to show that the limit satisfies a Nash fixed point equation.  Their more general approach uses  the Nash Equivalence Certainty principle introduced in \cite{caines1}. The limit equilibrium could or could not be a global optimal.
Here, we consider the general case where the dynamics and the cost may be arbitrary (we do not assume LQG Dynamics) so that  the optimal policy is not given in  closed form. The main  difference with their approach comes from the fact that we focus instead on centralized control to achieve a global optimum. The techniques to prove convergence are  rather different. Our proofs are more in line with classic mean field arguments  and  use stochastic approximation techniques.

Another example is  the work of Tembin\'e and others \cite{Tembine,Tembine2}, on the limits of games with many players. The authors  provide conditions under which  the limit when the number of players grows to infinity commutes  with the fixed point equation satisfied by a Nash equilibrium. Again, our investigation solves a different problem and focuses on  the centralized case. In addition, our approach is more algorithmic; we construct two intermediate systems: one with a finite number of objects controlled by a  limit policy and one with a limit system controlled by a stochastic policy induced by  the finite system.



\subsection*{Structure of the paper}

The rest of the paper is structured as follows. In
Section~\ref{sec:definition} we give definitions, some
notation and hypotheses. In Section~\ref{sec:mf_conv} we describe our main
theoretical.
In Section~\ref{sec-appli} we describe our resulting algorithm and illustrate the application of our
method with a few examples. The details of all proofs are in \sref{sec-proofs} and \sref{sec-conc} concludes the paper. 
\section{Notations and Definitions}
\label{sec:definition}

\subsection{System with $N$ Objects }

We consider a system
composed of  $N$ {\it objects}. Each object has a state from  the
finite set $\SS=\{1\dots S\}$. Time is discrete and the state
of the object $n$ at step $k\in\N$ is denoted $X^N_n(k)$. The
state of the system at time $k$ is $X^N(k)\bydef \lp
X^N_1(k)\dots X^N_N(k)\rp$.  For all $i\in\SS$, we denote by
$M^N(k)$ the empirical measure of the objects $\lp
X^N_1(k)\dots X^N_N(k)\rp$ at time $k$:
\begin{equation}
  M^N(k) \bydef \frac{1}{N}\sum_{n=1}^N \delta_{X^N_n(k)}, \label{eq:M^N(k)}
\end{equation}

where $\delta_x$ denotes the Dirac measure in $x$. $M^N(k)$ is
a probability measure on $\SS$ and its $i$th component $M^N(k)[i]$ denotes the
proportions of objects in state $i$ at time $k$ (also called
the occupancy measure): $M^N(k)[i]=\frac{1}{N}\sum_{n=1}^N \mathbf{1}_{X^N_n(k)=i}$.

The system $\lp X^N(k)\rp_{k \in \Nats}$ is a Markov process
once the sequence of the actions taken by the controller is
fixed. Let $\Gamma^N$ be the transition kernel, namely $\Gamma^N$ is a mapping $\SS^N\times\SS^N\times \A \to [0,1]$, where
$\A$ is the set of possible actions, such that for every $x\in \SS^N$ and $a \in \A$, $\Gamma^N(x,.,a)$ is a probability distribution on $\SS^N$ and further,
if the controller takes the
action $A^N(k)$ at time $t$ and the system is in state
$X^N(k)$, then:
 \be \PP\lp X^N(k+1)=y_1\dots
y_N|X^N(k)=x_1\dots x_N,A^N(k)=a\rp=\Gamma^N\lp x_1\dots
x_N,y_1\dots y_N,a\rp
 \label{eq-def-kernel}
 \ee
We assume that
\begin{quote}($A0$) Objects are observable only through their states
\end{quote}
in particular, the controller can observe the collection of all states $X^N_1,X_2^N,...$, but not the identities $n=1,2,...$. This assumption is required for mean field convergence to occur. In practice, it means that we need to put into the object state any information that is relevant to the description of the system.

Assumption ($A0$) translates into the requirement that the kernel be invariant by object re-labeling. Formally, let $\mathfrak{S}^N$ be the set of permutations of $\lc 1,2,...,N\rc$. By a slight abuse of notation, for $\sigma \in \mathfrak{S}^N$ and $x \in \SS^N$ we also denote with $\sigma(x)$ the collection of object states after the permutation, i.e.
$\sigma(x)\eqdef \lp x_{\sigma^{-1}(1)}...x_{\sigma^{-1}(N)}\rp$. The requirement is that
 \be
 \Gamma^N(\sigma(x),\sigma(y),a)=\Gamma^N(x,y,a)
 \ee for all $x,y \in \SS^N$, $\sigma\in \mathfrak{S}^N$ and $a \in \A$.
A direct consequence, shown in \sref{sec-proofs}, is:
\begin{theorem}For any given sequence of actions, the process $M^N(t)$ is a Markov chain
\label{theo-m-is-markov}
\end{theorem}

\subsection{Action, Reward and Policy}
\label{ssec:ARP}

At every time $k$, a centralized controller chooses an action
$A^N(k)\in\A$ where $\A$ is called the action set. $(\A,d)$ is
a compact metric space for some distance $d$. The purpose  of
Markov decision control is to compute optimal {\it policies}. A policy
$\pi=(\pi_0,\pi_1,\dots,\pi_k,\dots)$ is a sequence of decision rules that specify the action at every time instant. The policy $\pi_k$ might depend on the sequence of past and present
states of the process $X^N$, however, it
it known that when the state space is finite,  the action set compact and the kernel and the reward are continuous, there exists a deterministic
Markovian policy which is optimal (see Theorem 4.4.3 in \cite{puterman1994markov}). This implies that we can limit ourselves to policies that depend only on the current state $X^N(k)$.

Further, we assume that the controller can only observe object states. Therefore she cannot make a difference between  states that result from object relabeling, i.e. the policy depends on $X^N(k)$ in a way that is invariant by permutation. By \lref{lem-invariance} in \sref{sec-markov-om}, it depends on $M^N(k)$ only. Thus, we may assume that, for every $k$, $\pi_k$ is  a function $\PP(\SS)\to\A$. Let
$M^N_\pi(k)$ denotes the occupancy measure of the system at time $k$
when the controller applies policy $\pi$.

If the system has occupancy measure $M^N(k)$ at time $k$ and if the
controller chooses the action $A^N(k)$, she gets an
\emph{instantaneous reward} $r^N( M^N(k),A^N(k))$. The expected
 value over a finite-time horizon $[0;H^N]$ starting from
$m_0$ when applying the policy $\pi$ is defined by
\begin{equation}
  V^N_\pi(m) \bydef \espc{\sum_{k=0}^{\lfloor H^N \rfloor} r^N\lp M^N_\pi(k),\pi(M^N_\pi(k))\rp}
  {M^N_\pi(0)=m}
  \label{eq:reward_sto}
\end{equation}

The goal of the controller is to find an optimal policy that
maximizes the expected value. We denote by $V^N_*(m)$ the
optimal value when starting from $m$:
\be V^N_*(m) = \sup_{\pi} V^N_\pi(m)\label{eq-dev-VNstar}\ee


\subsection{Scaling Assumptions}

If at some time $k$, the system has occupancy measure $M^N(k)=m$ and the
controller chooses action $A^N(k)=a$, the system goes into
state $M^N(k+1)$ with probabilities given by the kernel
$Q^N(M^N(k),A^N(k))$. The expectation of the difference
between $M^N(k+1)$ and $M^N(k)$ is called the {\it drift} and
is denoted by $F^N(m,a)$:
\begin{equation}
 F^N\lp m,a\rp \bydef \E\lb M^N(k+1)-M^N(k) |
 M^N(k)=m,A^N(k)=a\rb.\label{eq:drift_FN}
\end{equation}
In order
to study the limit with $N$, we assume that $F^N$ goes to $0$
at speed $I(N)$ when $N$ goes to infinity and that $F^N/I(N)$
converges to a Lipschitz continuous function $f$. More
precisely, we assume that there exists a sequence
$I(N)\in(0;1)$, $N=1,2,3...$, called the \emph{intensity} of
the model with $\lim_{N\to\infty} I(N)=0$ and a sequence
$I_0(N)$, $N=1,2,3...$, also with $\lim_{N\to\infty} I_0(N)=0$
such that for all $m\in\PP(\SS)$ and $a\in\A$:
$\abs{\frac{1}{I(N)}F^N(m,a)-f(m,a)} \le I_0(N)$. In a sense,
$I(N)$ represents the order of magnitude of the number of
objects that change their state within  one unit of time.

The change of $M^N(k)$ during a time step is of order $I(N)$.
This suggests a rescaling of  time  by $I(N)$ to obtain an
asymptotic result. We define the continuous time process
$\lp\hat{M}^N(t)\rp_{t \in \Reals^+}$ as the affine
interpolation of $M^N(k)$, rescaled by the intensity function,
i.e. $\hat{M}^N$ is affine on the intervals $\lb k I(N), (k+1)
I(N)\rb$, $k \in \Nats$ and
\begin{equation*}
  \hat{M}^N ( k I(N) ) =  M^N(k).
\end{equation*}
Similarly, $\hat{M}^N_\pi$ denotes the affine interpolation of
the occupancy measure under policy  $\pi$. Thus, $I(N)$ can also be
interpreted as the duration of the time slot for the system
with $N$ objects.

We assume that the time horizon and the reward per time slot
scale accordingly, i.e. we impose
 \bearn
  H^N&=&\left\lfloor \frac{T}{I(N)} \right\rfloor\\
  r^N(m,a) & = & I(N) r(m,a)
  \eearn for every $m \in \PP(\SS)$ and $a \in \A$ (where $\lfloor x \rfloor$ denotes the largest integer $\leq x$).

\subsection{Limiting System (Mean Field Limit)}

We will see in Section~\ref{sec:mf_conv} that as $N$ grows, the
stochastic system $\hat{M}^N_\pi$ converges to a deterministic
limit $m_{\pi}$, the mean field limit. For more clarity, all
the stochastic variables (\emph{i.e.}, when $N$ is finite) are
in uppercase and their limiting deterministic values are in
lowercase.

An action function $\alpha:[0;T]\to \A$ is a piecewise
Lipschitz continuous function that associates to each time $t$
an action $\alpha(t)$. Note that action functions and policies
are different in the sense that action functions do not take
into account the state to determine the next action. For an action
function $\alpha$ and an initial condition $m_0$, we consider
the following ordinary integral equation for $m(t)$, $t\in
\Reals^+$:
\begin{equation}
  \label{eq:ode}
  m(t) - m(0)= \int_0^t f(m(s),\alpha(s))ds.
\end{equation}
(This equation is equivalent to an ODE, but is easier to manipulate in integral form. In the rest of the paper, we make a slight abuse of language and refer to it as an ODE). Under the foregoing assumptions on $f$ and $\alpha$, this equation
satisfies the Cauchy Lipschitz condition and therefore has a
unique solution once the initial condition $m(0)=m_0$ is fixed.
We call $\phi_t$, $t\in \Reals^+$, the corresponding semi-flow,
i.e.
\begin{equation}
  m(t) =\phi_t(m_0,\alpha) \label{eq:def_phit}
\end{equation}
is the unique solution of \eref{eq:ode}.



As for the system with $N$ objects, we define $v_\alpha(m_0)$
as the value of the limiting system over a finite horizon
$[0;T]$ when applying the action function $\alpha$ and starting
from $m(0)=m_0$:
\begin{equation}
  \label{eq:reward_deter}
  v_{\alpha}(m_0) \bydef \int_0^Tr\lp\phi_s(m_0,\alpha),\alpha(s)\rp ds.
\end{equation}
This equation looks  similar to the stochastic case
\eqref{eq:reward_sto} although there are two main differences.
The first is that  the system is deterministic. The second
is that it is defined for action functions and not for
policies. We also define the optimal value of the
deterministic limit $v_*(m_0)$:
\begin{equation}
  v_*(m_0) = \sup_{\alpha} v_\alpha(m_0)\label{eq-dev-vstar},
\end{equation}
where the supremum is taken over all possible action functions from
$[0;T]\to \A$.





\subsection{Table of Notations}

We recall here a list of the main notations  used throughout the paper. 

\newcommand\notation[2]{#1\dotfill #2\\}

\noindent
\notation{$M^N_\pi(k)$}{Empirical measure of the system with $N$ objects, under $\pi$, at time $k$, (Section \ref{ssec:ARP})} 
\notation{$F^N(m,a)$}{Drift of the system with $N$ objects when the state
  is $m$ and the action is $a$, \eref{eq:drift_FN}}
\notation{$f(m,a)$}{Drift of the limiting system (limit of rescaled $F^N(m,a)$ as
  $N \rightarrow \infty$), \eref{eq:drift_f}}
\notation{$\Phi_t(m_0,\alpha)$}{State of the limiting system: 
  $\Phi_t(m_0,\alpha)=m_0+\int_0^t f(\Phi_s(m_0,\alpha),\alpha(s))ds.$, \eref{eq:def_phit}}
\notation{$\pi^N$}{Policy for the system with $N$ objects: associates an
  action $a\in\A$ to each $k,M^N(k)$}
\notation{$\alpha$}{Action function for the limiting system: associates an
  action to each $t$: $\alpha:[0;T]\to\A$} 
\notation{$\pi^N_*$}{Optimal policy for the system with $N$ objects}
\notation{$\alpha_*$}{Optimal action function for the limiting system (if
  it exists)} 
\notation{$V^N_{\pi}(m)$}{Expected reward for the system with $N$ objects
  starting from $m$ under policy $\pi$, \eref{eq:reward_sto}} 
\notation{$V^N_*(m)$}{Optimal expected value for the system $N$:
  $V^N_*(m)=\sup_{\pi}V^N_\pi(m)=V^N_{\pi^*}(m)$, \eref{eq-dev-VNstar}} 
\notation{$V^N_{\alpha}(m)$}{Expected value for the system $N$
  when applying the action function $\alpha$, \eref{eq:reward_sto-2}} 
\notation{$v_\alpha(m)$}{Value of the limiting system starting from $m$
  under action function $\alpha$, \eref{eq:reward_deter}} 
\notation{$v_*(m)$}{Optimal value of the limiting system:
  $v_*(m)=\sup_\alpha v_\alpha(m)=v_{\alpha^*}(m)$, \eref{eq-dev-vstar}}


\subsection{Summary of Assumptions}
\label{sec:assum}

In \sref{sec:mf_conv} we establish theorems for the convergence
of the discrete stochastic optimization problem to a continuous
deterministic one. These theorems are based on several technical
assumptions, which are given next. Since $\SS$ is finite, the
set $\PP(\SS)$ is the simplex in $\Reals^{\SS}$ and for $m,m'
\in \PP(\SS)$ we define $\norm{m}$ as the $\ell^2$-norm of $m$
and $\cro{m,m'}=\sum_{i=1}^S m_i m'_{i}$ as the usual inner
product.
%
%

\paragraph{(A1) (Transition probabilities)}
Objects can be observed only through their
      state, \emph{i.e.}, the transition probability matrix (or  transition kernel)
      $\Gamma^N$, defined by \eref{eq-def-kernel}, is
      invariant under permutations of $1\dots N$.

There exist  some non-random functions $I_1(N)$
      and $I_2(N)$ such that $\lim_{N \to
      \infty}I_1(N)=\lim_{N \to \infty}I_2(N)=0$ and
      such that for all $m$ and any policy $\pi$, the
      number of objects that perform a transition
      between time slot $k$ and $k+1$ per time slot
      $\Delta^N_{\pi}(k)$ satisfies
    \begin{eqnarray*}
      \espc{\Delta^N_{\pi}(k)}{M^{N}_\pi(k)=m} & \leq &  N I_1(N)
      \\
      \espc{\Delta^N_{\pi}(k)^2}{M^{N}_\pi(k)=m} & \leq &  N^2 I(N) I_2(N)
    \end{eqnarray*}
    where $I(N)$ is the intensity function of the model, defined in the
    following assumption A2.

\paragraph{(A2) (Convergence of the Drift)} There exist
some non-random
  functions $I(N)$ and $I_0(N)$ and a function  $f(m,a)$ such that $\lim_{N \to
    \infty}I(N)=\lim_{N \to \infty}I_0(N)=0$ and
  \begin{equation}
   \norm{\frac{1}{I(N)}F^N(m,a)-f(m,a)} \leq I_0(N)
   \label{eq:drift_f}
  \end{equation}
  $f$ is defined on 
  $\PP(\SS)\times\A$ and there
  exists $L_2$ such that $\abs{f(m,a)}\le L_2$.

\paragraph{(A3) (Lipschitz Continuity)} There exist constants
    $L_1$, $K$ and $K_r$ such that for all
    $m,m'\in\PP(\SS)$, $a,a'\in\A$:
 \bearn
  \norm{F^N(m,a)-F^N(m',a)}&\leq& L_1 \norm{m-m'} I(N)\\
  \norm{f(m,a)-f(m',a')}&\leq &K(\norm{m-m'}+d(a,a'))\\
  \abs{r(m,a)-r(m',a)} & \leq & K_{r} \norm{m-m'} \eearn
We also assume that the reward is bounded: $\sup_{m,a\in
\A}
  \abs{r(m,a)} \bydef \norm{r}_{\infty} < \infty$.

To make things more concrete, here is a simple but useful case where all assumptions are true.
 \begin{itemize}
    \item There are constants $c_1$ and $c_2$ such that the expectation of the number of objects that perform a
transition in one time slot is $\leq c_1 $ and its standard deviation is $\leq c_2 $,
\item and $F^N(m,a)$ can be written under the form
    $\frac{1}{N}\varphi \lp m,a,1/N \rp$ where $\varphi$ is
    a continuous function on
  $\Delta_S \times \A \times [0, \epsilon)$ for some
  neighborhood $\Delta_S$ of $\PP(\SS)$ and some $\epsilon
  >0$, continuously differentiable with respect to $m$.
 \end{itemize}
In this  case we can choose $I(N)=1/N$, $I_0(N)= c_0/N$ (where
$c_0$ is an upper bound to the norm of the
differential $\pder{\varphi}{m}$),  $I_1(N)=c_1/N$
and $I_2(N)=(c_1^2 + c_2^2)/N$. 
\section{Mean Field Convergence}
\label{sec:mf_conv}

%
In \sref{sec-mfc-1} we establish the main results, then, in \sref{sec-mfc-2}, we provide the details of the method used to derive them.
\subsection{Main Results}
\label{sec-mfc-1}
The first result establishes convergence of the
optimization problem for the system with $N$ objects to the
optimization problem of the mean field limit:
\begin{theorem}[Optimal System Convergence] \label{th:exchange_limit}
Assume~(A0) to (A3). If $\lim_{N\to\infty} M^N(0) = m_0$ almost
surely [resp. in probability] then:
  \begin{equation*}
    \lim_{N\to\infty} V^{N}_*\lp M^N(0)\rp = v_*\lp m_0\rp
  \end{equation*}
almost surely [resp. in probability], where $V^N_*$ and $v_*$ are the optimal values for the system with $N$ objects and the mean field limit, defined in \sref{sec:definition}.
\end{theorem} The proof is given in \sref{sec-proof-main}.

The second result states that an optimal action function for the mean field
limit provides an asymptotically optimal strategy for the system with $N$
objects. We need, at this point, to introduce a first auxiliary system, which
is a system with $N$ objects controlled by an action function borrowed from
the mean field limit. More precisely, let $\alpha$ be an action function
that specifies the action to be taken at time $t$. Although $\alpha$ has
been defined for the limiting system, it can also be used in the system
with $N$ objects. In this case, the action function $\alpha$ can be seen as
a policy that does not depend on the state of the system. At step $k$, the
controller applies action $\alpha(kI(N))$. By abuse of notation, we denote
by $M^N_\alpha$, the state of the system when applying the action function
$\alpha$ (it will be clear from the notation whether the subscript is an
action function or a policy). The value for this system is defined by
\begin{equation}
  V^N_\alpha(m_0) \bydef \espc{\sum_{k=0}^{H^N} r\lp M^N_\alpha(k),\alpha(kI(N))\rp}
  {M^N_\alpha(0)=m_0}
  \label{eq:reward_sto-2}
\end{equation}
Our next result is the convergence of
convergence of $M^N_\alpha$ and of the value:
\begin{theorem}
  \label{theo:conv_reward1}
Assume (A0) to (A3); $\alpha$ is a piecewise Lipschitz continuous
action function on $[0;T]$, of constant $K_\alpha$, and with at
most $p$
  discontinuity points. Let $\hat{M}^N_{\alpha}(t)$ be the
  linear interpolation of the discrete time process
  $M^N_\alpha$.
  Then for all $\epsilon>0$: 
 \be
 \P
    \lc
    \sup_{0\leq t \leq T}\norm{\hat{M}^N_{\alpha}(t)-\phi_t(m_0,\alpha)}
    >
    \lb
    \norm{M^N(0)-m_0}  + I_0'(N,\alpha) T+ \eps
    \rb
    e^{L_1 T}
    \rc
    \leq
    \frac{ J(N,T)}{\eps^2}\label{eq-q-conv-p2}
 \ee
and
\be
  \abs{
      V^{N}_{\alpha}\lp M^N(0)\rp - v_{\alpha}(m_0)
     }
     \leq
     B'\lp
      N , \norm{M^N(0)-m_0}\rp \label{eq-q-conv-p-v2}
  \ee
  where $J, I'_0$ and $B'$ are defined in \sref{sec-def-cts}
  and satisfy $\lim_{N\to\infty}I'_0(N,\alpha)=\lim_{N\to\infty}J(N,T)=0$ and $\lim_{N\to\infty,\delta\to 0} B'(N,\delta)=0$.

 In particular, if $\lim_{N\to\infty} M^N_\pi(0) = m_0$ almost surely [resp. in probability]  then $\lim_{N \to \infty}
      V^{N}_{\alpha}\lp M^N(0)\rp=  v_{\alpha}(m_0) $
     almost surely [resp. in probability].

%
\end{theorem}
The proof is given in \sref{sec:coro1_proof}.

As the reward function
$r(m,a)$ is bounded and the time-horizon $[0;T]$ is finite, the
set of values when starting from the initial condition $m$,
$\{v_\alpha(m):\alpha \mathrm{~action~function}\}$, is bounded.
This set is not necessarily compact because the set of action
functions may not be closed (a limit of Lipschitz continuous
functions is not necessarily Lipschitz continuous). However, as
it is bounded, for all $\epsilon>0$, there exists an action
function $\alpha^\epsilon$ such that $v_*(m) = \sup_\alpha
v_\alpha(m) \le v_{\alpha^\epsilon}+\epsilon$.
Theorem~\ref{th:exchange_limit} shows that $\alpha^\epsilon$ is
optimal up to $2\epsilon$ for $N$ large enough.
This shows the following corollary:
\begin{coro}[Asymptotically Optimal Policy]Let $\alpha^*$ be an optimal action
function for the limiting system. Then
\ben\lim_{N\to \infty}\abs{V^N_{\alpha^*}-V^{N}_*}=0
\een
In other words, an optimal action function for the
limiting system is asymptotically optimal for the system with
$N$ objects.
\label{coro-as-opt}
\end{coro}

In particular, this shows that as $N$ grows, policies that do
not take into account the state of the system (\emph{i.e.},
action functions) are asymptotically as good as adaptive
policies. In practice however, adaptive policies might perform
better, especially for very small values of $N$. 
However, it is in general impossible to prove convergence for
 adaptive policies.


\subsection{Derivation of Main Results}\label{sec-mfc-2}

\subsubsection{Second Auxiliary System}
The method of proof uses a second auxiliary system, the process
$\phi_t(m_0,A^N_\pi)$ defined below. It is a limiting system controlled by an action function derived from the policy of the original system with $N$
objects.

Consider the system with $N$ objects under policy $\pi$. The
process $M^N_\pi$ is defined on some probability space
$\Omega$. To each $\omega\in\Omega$ corresponds a trajectory
$M^N_\pi(\omega)$, and for  each $\omega\in\Omega $, we define an
action function $A^N_\pi(\omega)$. This  random function is
piecewise constant on each interval $[kI(N),(k+1)I(N))$
($k\in\N$) and is  such that $A^N_{\pi}(\omega)(kI(N))\bydef
\pi_k(M^N(k))$ is the action taken by the controller of the
system with $N$ objects at time slot $k$, under policy $\pi$.

Recall that for any $m_0\in\PP(\SS)$ and any action function
$\alpha$, $\phi_t(m_0,\alpha)$ is the solution of the
ODE~\eqref{eq:ode}. For every $\omega$,
$\phi_t(m_0,A^N_{\pi}(\omega))$ is the solution of the limiting
system with action function $A^N_{\pi}(\omega)$, i.e.
 \ben
 \phi_t(m_0,A^N_{\pi}(\omega)) - m_0= \int_0^t f(\phi_s(m_0,A^N_{\pi}(\omega)),A^N_{\pi}(\omega)(s))ds.
 \een

%
When $\omega$ is fixed, $\phi_t(m_0,A^N_{\pi}(\omega))$ is a continuous time
deterministic process corresponding to one trajectory
$M^N_\pi(\omega)$. When considering all possible realizations of $M^N_\pi$,
$\phi_t(m_0,A^N_{\pi})$ is a random, continuous time function ``coupled'' to
$M^N_{\pi}$. Its randomness comes only from the action term
$A^N_{\pi}$,  in the
ODE. In the following, we omit to write the dependence in
$\omega$. $A^N_\pi$ and $M^N_\pi$ will always designate the processes
corresponding to the same $\omega$.

\subsubsection{Convergence of Controlled System}The following result is the main technical result; it shows the
convergence of the controlled system in probability, with
explicit bounds.  Notice that it does not require any
regularity assumption on the policy~$\pi$.
\begin{theorem}
  \label{theo-bounds}
  Under Assumptions (A0) to (A3), for any $\eps >0$, $N \geq 1$
  and any policy $\pi$:
  \begin{equation}
    \P
    \lc
    \sup_{0\leq t \leq T}\norm{\hat{M}^N_{\pi}(t)-\phi_t(m_0,A^N_{\pi})}
    >
    \lb
    \norm{M^N(0)-m_0}  + I_0(N) T+ \eps
    \rb
    e^{L_1 T}
    \rc
    \leq
    \frac{ J(N,T)}{\eps^2}
    \label{eq:theo-bounds}
  \end{equation}
  where $\hat{M}^N_{\pi}$ is the linear interpolation of the discrete
time system with $N$ objects) and $J$ is defined in \sref{sec-def-cts}.
\end{theorem}
Recall that $I_0(N)$ and $J(N,T)$ for a fixed $T$  go to $0$ as $N\to
\infty$. The proof is given in \sref{sec-proof-maintheo}.

\subsubsection{Convergence of Value}
Let $\pi$ be a policy and $A^N_\pi$ the sequence of actions
corresponding to a trajectory $M^N_\pi$ as we just defined.
\eref{eq:reward_deter} defines the value for the deterministic
limit when applying a sequence of actions. This defines a random
variable $v_{A^N_\pi}(m_0)$ that corresponds to the value over the
limit system  when using $A^N_\pi$ as action function. The random part comes
from $A^N_\pi$. $\E\lb v_{A^N_{\pi}}(m_0)\rb$ designates the
expectation of this value over all possible $A^N_\pi$. A first
consequence of Theorem~\ref{theo-bounds} is the convergence of
$V^{N}_{\pi}\lp M^N(0)\rp$ to $\E\lb v_{A^N_{\pi}}(m_0)\rb$
with an error that can be uniformly bounded.

\begin{theorem}[Uniform convergence of the value]
  \label{th:conv_reward2}
  Let $A^N_{\pi}$ be the random action function associated with
  $M^N_{\pi}$, as defined earlier.
  Under Assumptions (A0) to (A3),
  \ben
  \abs{
      V^{N}_{\pi}\lp M^N(0)\rp - \E\lb v_{A^N_{\pi}}(m_0) \rb
     }
     \leq
     B\lp
      N , \norm{M^N(0)-m_0}\rp
  \een
where $B$ is defined in \sref{sec-def-cts}.

Note that $\lim_{N\to\infty,\delta\to 0} B(N,\delta)=0$;
 in particular, if $\lim_{N\to\infty} M^N_\pi(0) = m_0$ almost surely [resp. in probability]  then $\abs{
      V^{N}_{\pi}\lp M^N(0)\rp - \E\lb v_{A^N_{\pi}}(m_0) \rb
     }\to 0$ almost surely [resp. in probability].
 \end{theorem}


The proof is given in \sref{proof:conv_reward}.

\subsubsection{Putting Things Together}The proof of the main result uses the two auxiliary systems. The first auxiliary system provides a strategy for the system with $N$ objects derived from an action function of the mean field limit; it cannot do better than the optimal value for the system with $N$ objects, and is close to the optimal value of the mean field limit. Therefore, the optimal value for the system with $N$ objects is  lower bounded by the optimal value for the mean field limit. The second auxiliary system is used in the opposite direction, which shows that, roughly speaking, for large $N$ the two optimal values are the same. We give the details of the derivation in \sref{sec-proof-main}.


\section{Applications}
\label{sec-appli}
\subsection{Hamilton-Jacobi-Bellman Equation and Dynamic Programming}

Let us now consider the finite time optimization problem for the stochastic
system and its limit from a constructive point of view.
As the state space is finite, we can compute the
optimal value by using a dynamic programming algorithm.
If $U^N(m,t)$ denotes the optimal value for the stochastic system
starting from $m$ at time $t/I(N)$, then  $U^N(m,t)=\sup_\pi \E\lb
\sum_{k=t/I(N)}^{T/I(N)} r^N( M^N_\pi(k) ) : M^N(t) = m
\rb$.
The optimal value can
be computed by a discrete dynamic programming algorithm
\cite{puterman1994markov} by
setting $U^N(m,T)=r^N(m)$ and
\begin{equation}
\label{dp}
  U^N(m,t) = \sup_{a\in\A} \espc{r^N(m,a) + U^N(M^N(t+I(N)),
    t+I(N))}{\bar{M}^N(t) = m, A^N(t)=a}.
\end{equation}
Then, the  optimal cost over horizon $[0;T/I(N)]$ is $V^N_*(m)=U(m,0)$.

Similarly, if we denote by $u(m,t)$ the optimal cost over
horizon $[t;T]$ for the limiting system, $u(m,t)$ satisfies the
classical Hamilton-Jacobi-Bellman equation:
\begin{equation}
\label{hjb}
\dot{u}(m,t) + \max_a \left\{ \nabla u(m,t).f(m, a) + r(m,a) \right\} =
0.
\end{equation}
This  provides a way to compute the optimal value, as well as
the optimal policy,  by solving the
partial differential equation above.

\subsection{Algorithms }

Theorem \ref{th:exchange_limit} above can be used to design  an
effective construction of an asymptotically optimal policy for
the system with $N$ objects over the horizon $[0,H]$ by using
the procedure described in Algorithm~\ref{algo-stat}.

\begin{algorithm}
 \Begin{From the original system with $N$ objects, construct
    the occupancy measure $M^N$ and its kernel $\Gamma^N$ and
    let $M^N(0)$ be the initial occupancy measure\;

  Compute the limit of the drift of $\Gamma^N$, namely
    the function $f$\;

    Solve the HJB equation \eqref{hjb} on the interval $[0,HI(N)]$.
This provides an optimal
  control function $\alpha(M^N_0,t)$\;

Construct a discrete control $\pi(M^N(k),k)$ for the discrete
system, that gives the action to be taken under state $M^N(k)$
at step $k$:
\[ \pi(M^N(k), k) \bydef  \alpha(\phi_{k I(N)}(M^N(0),\alpha)).\]

 \Return{$\pi$\;}}
 \label{algo-stat}
 \caption{Static algorithm constructing a
   policy for the system with N objects, over the finite horizon.}
\end{algorithm}
%
%
%
%

Theorem \ref{th:exchange_limit} says that under policy $\pi$, the total  value
$V_{\pi}^N $  is asymptotically optimal:
\begin{equation*}
    \lim_{N\to\infty}V^{N}_{\pi}(M^N(0)) =  \liminf_{N\to\infty}
    V^N_*(M^N(0)).
  \end{equation*}

The policy $\pi$ constructed by Algorithm~\ref{algo-stat} is
static in the sense that it does not depend on the state
$M^N(k)$ but only on the initial state $M^N(0)$, and the
deterministic estimation of $M^N(k)$ provided by the
differential equation. One can construct a more adaptive policy
by updating the starting point of the differential equation at
each step. This new procedure, constructing an adaptive policy
$\pi'$ from  0 to the final horizon $H$ is given in
Algorithm~\ref{adapt}.

 \begin{algorithm}
 \Begin{
$M:= M^N(0)$; $k:=0$\;
 \Repeat{k=H}
{$\alpha_k(M,\cdot) := $ solution of \eqref{hjb} over
  $[k I(N),H I(N)]$ starting in $M$\;
$\pi'(M, k) :=  \alpha_k(\phi_{k I(N)}(M,\alpha_k))$\; $M$ is
changed by applying  kernel $\Gamma^N_{\pi'}$\; k:= k+1\;}
\Return{$\pi'$\;}} \caption{Adaptive algorithm constructing a
policy $\pi'$ for the
  system with $N$ objects, over the finite horizon $H$.}
\label{adapt}
\end{algorithm}

In practice, the total value of the adaptive policy $\pi'$ is larger
than the value of the static policy $\pi$ because it uses on-line
corrections at  each step, before taking a new action. However Theorem
\ref{th:exchange_limit} does not provide a proof of its asymptotic optimality.

\subsection{Examples}
\label{sec:appli}

In this section, we develop three examples.
The first one can be seen as a simple illustration of
optimal mean field. The limiting ODE  is quite simple and can be
optimized in closed analytical  form.

The second example considers a classic virus problem.  Although virus
propagations concern discrete objects (individuals or devices), most work
in the literature study a continuous approximation of the problem under
the form of an ODE.  The justification of passing from a discrete to a
continuous model is barely mentioned in most papers (they mainly focus on
the study of the ODE).  Here we present a discrete dynamical system based
on a simple stochastic mobility model for the individuals whose behavior
converges to a classic continuous model. We show on a numerical example
that the limiting problem provides a policy that is close to optimal, even
for a system with a relatively small numbers of nodes.

Finally, the last example comes from routing optimization in a
queueing network model of volunteer computing platforms.
The purpose of this last example is to show that a discrete optimal
control problem suffering from the curse of dimensionality can be
replaced by a continuous optimization problem where an HJB equation
must be solved over a much smaller state space.


\subsubsection{Utility Provider Pricing}
\label{sec:harvest}

This is  a simplified
discrete Merton's problem. This  example
shows a case where the optimization problem in the infinite system can
be solved  in closed form. This can be seen as an ideal case
for  the mean field approach:
although the original system is difficult to solve even numerically when
$N$ is large,
taking the limit when $N$ goes to infinity makes it simple to solve,
in an  analytical form.

We consider a system made of a utility and $N$
users; users can be either in state $S$ (subscribed) or $U$ (unsubscribed). The utility fixes their price $\alpha\in [0,1]$. At every time step, one randomly chosen customer revises her status: if she is in state $U$ [resp. $S$], with probability $s(\alpha)$ [resp. $a(\alpha)$] she moves to the other state; $s(\alpha)$ is the probability of a new subscription, and $a(\alpha)$ is the probability of attrition. We assume $s(\cdot)$ decreases with $\alpha$ and $a(\cdot)$ increases. If the price is large, the instant gain is large, but the utility loses customers, which eventually reduces the gain.

\begin{figure}[h]
  \centering
  \includegraphics[width=10cm]{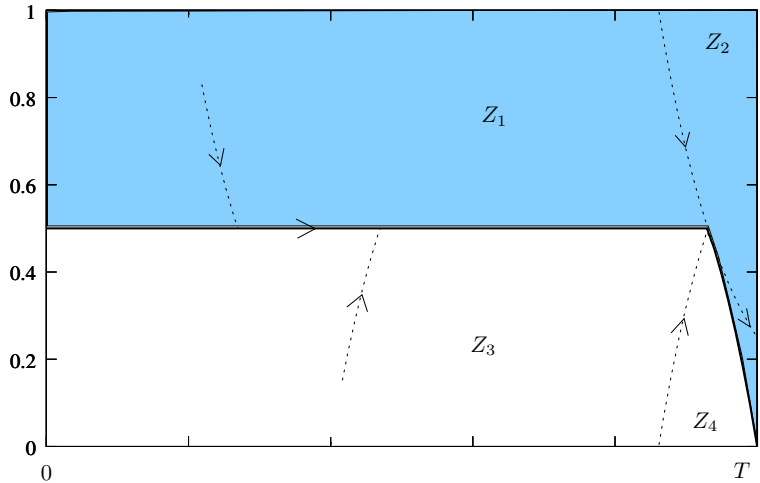}
  \caption{Evolution of the proportion of subscribers ($y$-axis) under the optimal pricing policy.}
  \label{fig:policyEx1}
\end{figure}

Within our framework, this problem can be seen as a Markovian  system
made of $N$ objects (users) and one controller (the provider). The intensity of
the model is $I(N)=1/N$.
Moreover, if the immediate profit  is divided by $N$ (this does not
alter the optimal pricing policy) and if $x(t)$ is the fraction of
objects in state $S$ at time $t$
and $\alpha(t)\in[0;1]$ is  the action taken by the provider at time
$t$, the mean field limit of the system is:
\begin{equation}
  \label{eq:harvest_diff_eq0}
  \pder{x}{t} = -x(t)a(\alpha(t)) + (1-x(t))s(\alpha(t)) = s(\alpha(t))- x (s(\alpha(t))+a(\alpha(t))
\end{equation}
and the rescaled profit over a time horizon $T$ is $\int_0^T x(t)\alpha(t)dt$. Call  $u_*(t,x)$ the optimal benefit over the interval
$[t,T]$ if there is a proportion $x$ of subscribers  at time $t$. The
Hamilton-Jaccobi-Bellman equation is
\bearn
\label{hjbEx1-0} \frac{\partial}{\partial t}u_*(t,x)+
 H\lp x,\frac{\partial}{\partial
  x}u_*(t,x)\rp &=&0
 \\
 \mbox{with } H(x,p) &=& \max_{\alpha \in [0,1]}\lb p(s(\alpha)- x (s(\alpha)+a(\alpha)) + \alpha x\rb
\eearn
$H$ can be computed under reasonable assumptions on the rates of subscription and attrition $s()$ and $a()$, which can then be used to show that there exists  an optimal policy that is threshold based. To continue the rest of this illustration, we consider the radically simplified case where $\alpha$ can take only the values $0$ and $1$ and under the  conditions $s(0) = a(1) =1$ and $s(1) = a(0) = 0$, in which case the ODE becomes
 \begin{equation}
  \label{eq:harvest_diff_eq}
  \pder{x}{t} = -x(t)\alpha(t) + (1-x(t))(1-\alpha(t)) = 1 - x(t) - \alpha(t),
\end{equation}
and $H(x,p)=\max\lp x(1-p),(1-x)p \rp$. The solution of the HJB equation can be given in closed form.
The optimal policy  is to chose action $\alpha=1$ if $x>1/2$ or $x>1-\exp(-(T-t))$, and
$0$ otherwise.
Figure \ref{fig:policyEx1} shows the evolution of the proportion of
subscribers $x(t)$ when the optimal policy is used.
The coloured area  corresponds to all the points $(t,x)$ where the optimal
policy is $\alpha = 1$ (fix a high price)
and the white area  is where the optimal policy is to choose $\alpha
= 0$ (low price).

To show that this policy is indeed optimal,
one has to compute the corresponding value of the benefit $u(t,x)$ and show that
it satisfies the HJB equation.
This can be done using a case analysis, by computing explicitly  the value of
$u(t,x)$  in the zones $Z_1,Z_2,Z_3$ and $Z_4$ displayed in Figure
\ref{fig:policyEx1},  and check
that $u(t,x)$ satisfies \eref{hjbEx1-0} in each case.

\subsubsection{Infection Strategy of a Viral Worm}
\label{sec:vaccination}


This second example has two purposes.  The first one is to provide a
rigorous justification of the use of a continuous optimization approach for
this classic problem in population dynamics and to show that the
continuous limit provides insights on the structure of the optimal behavior
for the discrete system. Here, the optimal action function can be shown to
be of the bang-bang type for the limit problem, by using tools from continuous
optimization such as the Pontryagin maximum principle. Theorem
\ref{th:exchange_limit} shows that a bang-bang policy should also be
asymptotically optimal in the discrete case.

The second purpose is to compare numerically the performance of the optimal
policy of the deterministic limit $\alpha_*$ and the performance of other
policies for the stochastic system for small values of $N$. We show that
$\alpha_*$ is close to optimal even for $N=10$ and that it outperforms
another classic heuristic.

This example is taken from \cite{Saswati} and considers the propagation of
infection by a viral worm. Actually, similar epidemic models have been
validated through experiments, as well as simulations as a realistic
representation of the spread of a virus in mobile wireless networks (see
\cite{Cole,Helmy}).  A {\it susceptible} node is a mobile wireless device,
not contaminated by the worm but prone to infection. A node is {\it
  infective} if it is contaminated by the worm. An infective node spreads
the worm to a susceptible node whenever they meet, with probability $\beta$.
The worm can also choose to kill an infective node, i.e., render it
completely dysfunctional - such nodes are denoted {\it dead}.  A functional
node that is immune to the worm is referred to as {\it recovered}. Although
the network operator uses security patches to immunize susceptibles (they
become recovered) and heals infectives to the recovered state, the goal of
the worm is to maximize the damages done to the network.  Let the total
number of nodes in the network be $N$. Let the proportion of susceptible,
infective, recovered and dead nodes at time $t$ be denoted by $S(t)$, $I
(t)$, $R(t)$ and $D(t)$, respectively.  Under a uniform mobility model, the
probability that a susceptible node becomes infected is $\beta I/N$.  The
immunization of susceptibles (resp. infectives) happens at a fixed rate $q$
(resp. $b$). This means that a susceptible (resp. infective) node is
immunized  with probability $q/N$ (resp. $b/N$) at every time step.

At this point, authors of \cite{Saswati} invoke the classic results of
Kurtz \cite{Kurtz70} to show that the dynamics of this population process
converges to the solution of the following differential equations.
\begin{equation}
  \label{eq:sir_diff_equation}
  \begin{array}{rcl}
    \pder{S}{t} &=& - \beta  IS -qS\\
    \pder{I}{t} &=& \beta  IS - bI - v(t) I\\
    \pder{D}{t} &=& v(t) I \\
    \pder{R}{t} &=& bI + qS.
  \end{array}
\end{equation}
This system actually satisfies assumptions ($A_1,A_2,A_3$), which allows us
not only to obtain the mean field limit, but also to say more about the
optimization problem.  The objective of the worm is to find $v(\cdot)$ such
that the damage function $D(T) + \int_{0}^T f(I(t))dt$ is maximized under
the constraint $0 \leq v \leq v_{\max} $ (where $f$ is convex).  In
\cite{Saswati}, this problem is shown to have a solution and the Pontryagin
maximum principle is used to show that the optimal solution $v_*(\cdot)$ is
of bang-bang type:
\begin{equation}
  \exists t_1 \in  [0 \ldots T ) \mbox{ s. t. } v_*(t) = 0
  \mbox{ for } 0 < t < t_1 \mbox{ and } v_*(t) = v_{\max} \mbox{ for } t_1
  < t < T.
  \label{eq:bang-bang}
\end{equation}

Theorem \ref{th:exchange_limit} makes the formal link between the
optimization of the model on an individual level and the previous
resolution of the optimization problem on the differential equations, done
in \cite{Saswati}.  It allows us to formally claim that the policy
$\alpha_*$ of the worm is indeed asymptotically optimal when the number of
objects goes to infinity.

\begin{figure}[ht]
  \centering
  \begin{tabular}{cc}
    \subfigure[\label{subfig:1}]{\includegraphics[width=.48\linewidth]{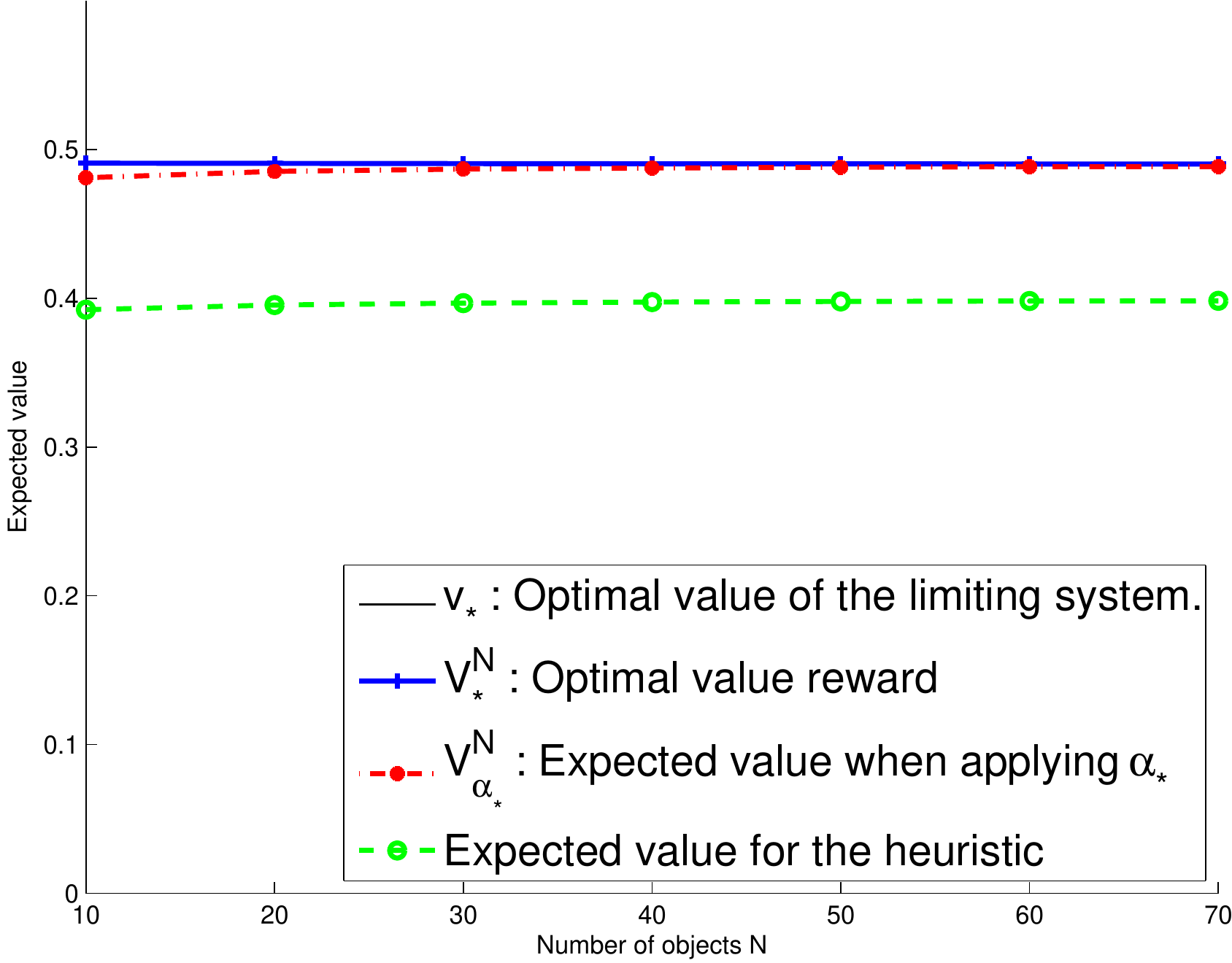}}~~&
    \subfigure[\label{subfig:zoom}
Same as (a) with $y-$axis zoomed around $0.49$]
{\includegraphics[width=.48\linewidth]{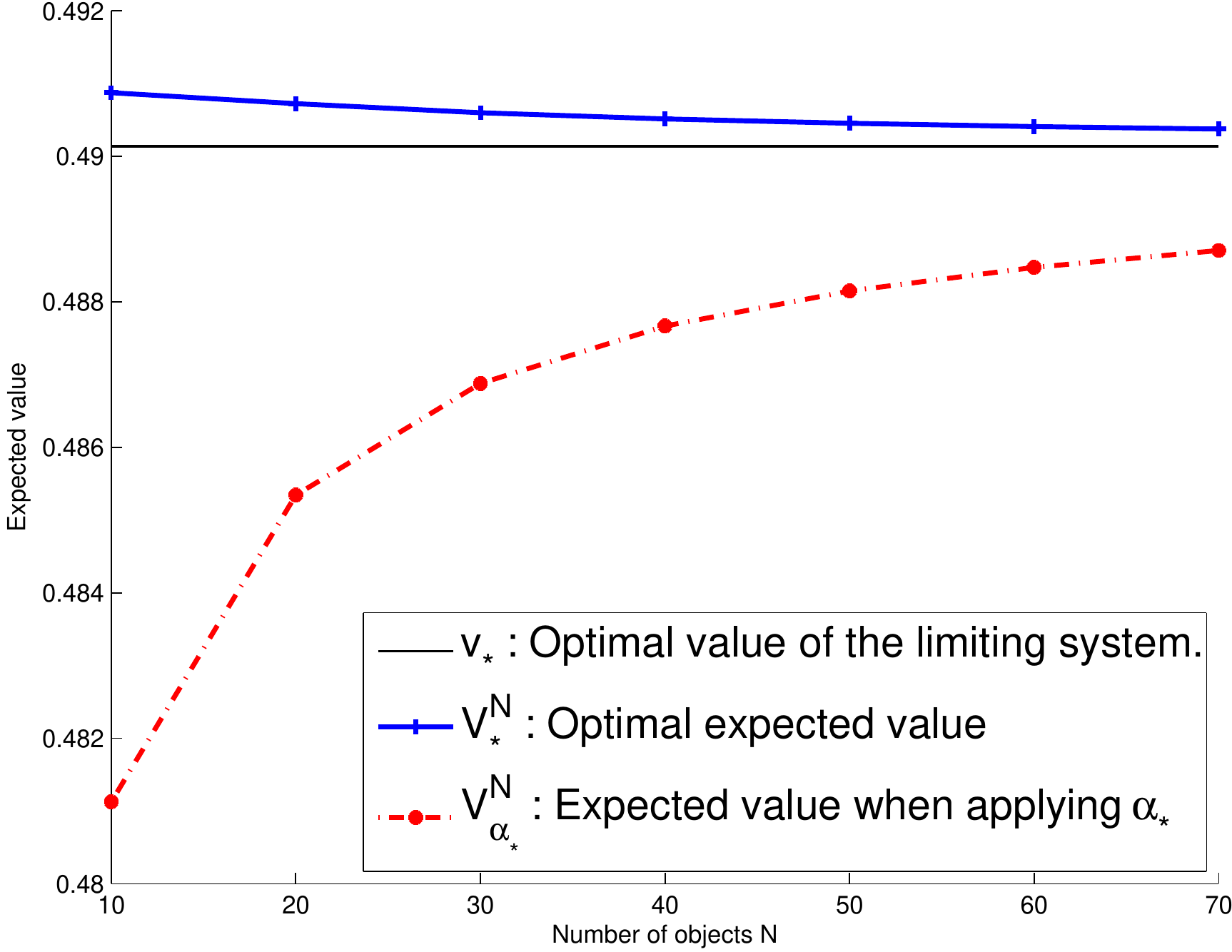}}
\end{tabular}
\caption{Damage caused by the worm for various infection policies as a
  function of the size of the system $N$. 
  The goal of the worm is to \emph{maximize} the damage (higher means better). Panel~(a) shows the optimal value $v_*$ for the limiting system (mean field limit), the optimal value $V^N_*$ for the system with $N$ objects, the value $v^N_{\alpha_*}$ of the asymptotically optimal policy given in \coref{coro-as-opt} and the value of a classic heuristic. Panel~(b) zooms the $y-$axis around the values of the optimal policies.
}
  \label{fig:virus}
\end{figure}

We investigated numerically the performance of $\alpha_*$ against various
infection policies for small values of the number of nodes in the system
$N$. These results are reported on Figure~\ref{fig:virus}, where we compare
four values:
\begin{itemize}
\item $v_*$ -- the optimal value of the limiting system.
\item $V^N_*$ -- the optimal expected damage for the system with $N$
  objects (MDP problem);
\item $V^N_{\alpha_*}$ -- the expected value of the system with $N$
  objects when applying the action function $\alpha_*$ that is optimal for
  the limiting system; Performance of algorithm~\ref{algo-stat}
\item the performance of a heuristic where, instead of choosing a
  threshold as suggested by the limiting system \eqref{eq:bang-bang}, the
  killing probability $\nu$ if fixed for the whole time. The curve on the
  figure is drawn for the optimal $\nu$ (recomputed for each parameter
  $N$).
\end{itemize}
We implemented a simulator that follows strictly the model of infection
described earlier in this part. We chose parameters similar to those used
in \cite{Saswati}: the parameter for the evolution of the system are
$\beta=.6$, $q=.1$, $b=.1$, $v_{\max{}}=1$ and the damage function to be
optimized is $D(T) + \frac1T\int_{0}^T I^2(t)dt$ with $T=10$. However, it
should be noted that the choice of thess parameters does not influence
qualitatively the results. Thanks to the relatively small size of the
system, these four quantities can be computed numerically using a backward
induction. The optimal policies for the deterministic limit consists in not killing
machines until $t_1=4.9$ and in killing machines at a maximum rate after that time:
$\alpha_*(t)=\mathbf{1}_{\{t>4.9\}}$.

Theorem~\ref{th:exchange_limit} shows that $\alpha_*$ is
asymptotically optimal ($\lim_{N\to\infty}V^N_* = \lim_{N\to\infty}
V^N_{\alpha_*} = v_*$), but Figure~\ref{subfig:1} shows that, already for low
values of $N$, these three quantities are very close. A classic
heuristic for this maximal infection problem is to kill a node with a
constant probability $\nu$, regardless of the time horizon. Our numerical
study shows that $\alpha_*$ outperforms this heuristic by more than
$20\%$.  The performance of this heuristic does not increase with the size of
the system $N$.

In order to illustrate the convergence of the values $V^N_*$ and
$V^N_{\alpha_*}$ to $v_*$, Figure~\ref{subfig:zoom} is a detailed view of
Figure~\ref{subfig:1} where we show the two quantities $V^N_*$,
$V^N_{\alpha_*}$ and their common limit $v_*$.  This figure shows that the
convergence is indeed very fast. Other numerical experiments indicate that
this is true for a large panel of parameters.  Although this figures seems
to indicate that $V^N_{\alpha_*}\le v_*\le V^N_*$, this is not true in
general, for example adding $5D(t)$ to the damage function leads to
$V^N_{\alpha_*}\le V^N_* \le v_*$ ($V^N_{\alpha_*}$ is always less than
$V^N_*$ by definition of $V^N_*$).

\subsubsection{Brokering Problem}
\label{sec:boinc}

Finally, let us consider a model of a volunteer computing system such as BOINC
\url{http://boinc.berkeley.edu/}. Volunteer computing means that people
make  their personal computer available for a computing system. When they do
not use their computer, it is available for the computing
system. However, as soon as they start using their computer, it becomes
unavailable for the computing system. These systems are becoming more and more
popular  and provide  large computing power at a very low cost
\cite{kondo}.

\begin{figure}[hbtp]
  \centering
  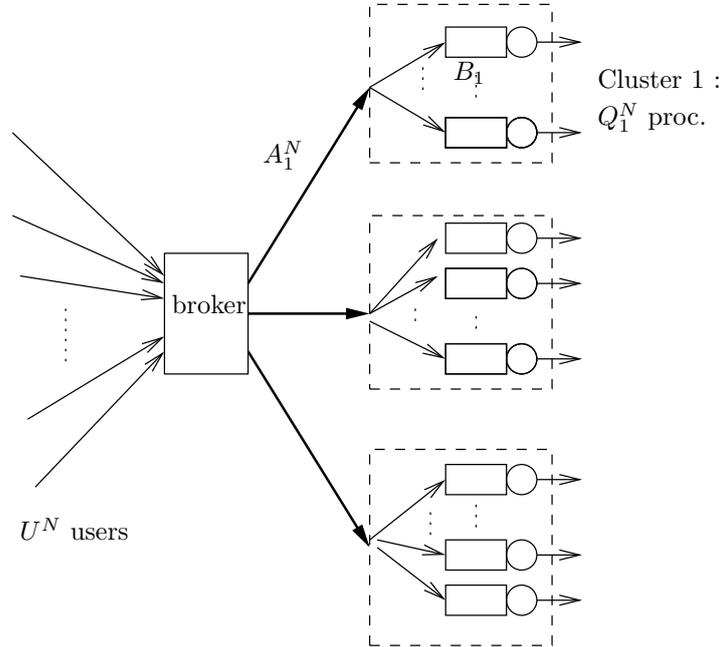
  \caption{The brokering problem in a desktop grid system, such as Boinc}
  \label{fig:boinc}
\end{figure}

The Markovian model  with $N$ objects is defined as follows. The $N$ objects
represent the  users that can submit jobs to the system and the resources that can run the jobs. The resources are
grouped into  a small number of clusters and all resources in the same
cluster share the same characteristics in terms of speed and
availability. Users send jobs to a central broker whose role is to balance
the load among the clusters.

The model is a discrete time model of a queuing system. Actually, a
more natural  continuous-time Markov model  could also be handled similarly, by using uniformization.

There are $U^N$ users. Each user has a state
$x \in\{\mathrm{on}, \mathrm{off}\}$.  At each time
step, an active user {\bf  s}ends  one  job with
probability $p_s^N$ and becomes {\bf i}nactive with probability
$p_i/N$. An inactive user sends no jobs to the system and becomes {\bf
  o}n  with probability $p_{o}/N$.

There are $C$ clusters in the system.
Each cluster $c$ contains $Q_c^N$ computing resources. Each
resource has a buffer of bounded size $J_{c}$. A resource can either be
valid or broken. If it is valid and if it has one or more job in its
queue, it completes one job with probability $\mu_c/N$ at this time
slot. A  resource gets {\bf b}roken with probability $p_b/N$. In that case, it discards all the packets of its buffer.  A broken resource becomes {\bf v}alid with probability $p_v/ N$.

At each time step, the broker takes an action $a\in \PP(\{1\dots C\})$ and
sends the packets it received to the clusters according to the distribution
$a$. A packet sent to cluster $c$ joins the queue of one of the resources, sqy $k$;
according to a local rule (for example chosen uniformly among the $Q_c^N$ resources composing the cluster).
If the queue of resource $k$ is full, the packet is
lost. The goal of the broker is to minimize  the number of losses plus the total size of the queues
over a finite horizon (and hence the response time of accepted packets).

This model is represented in Figure~\ref{fig:boinc}.

The system has  an intensity $I(N)\bydef 1/N$.  The number $C$
of clusters is fixed and does not depend on $N$, as well as the sizes $J_c$
of the buffers. However, both the number of users $U^N$,  and the number of
resources in the clusters $Q^N_c$,  are linear in  $N$. Finally, by
construction, all the state changes occur with probabilities that
scale with $1/N$.

The limiting system is described by the variable $m_o(t)$, that represents
the fraction of users who are on, and the variables $q_{c,j}(t)$ and $b_{c}(t)$
that, respectively, represent the fraction of resources in cluster $c$ having $j$ jobs in their buffer
and the fraction of resources in cluster $c$ that are broken. For an action
function $\alpha(\cdot)$, we denote by $\alpha_c(\cdot)$ the fraction of packets
sent to cluster $c$. Finally, let us denote  by $m$ the fraction of
users (both active or inactive) and $q_c$ the fraction of processors
in cluster $c$. These fractions are constant (independent of time) and satisfy $m + q_1+\cdots + q_C =1$. We get the
following equations:
\begin{eqnarray}
  \pder{m_o(t)}{t} &=& - p_i m_o(t) + p_o(m-m_o(t)) \label{eq:boinc_ode_0}\\
  \pder{q_{c,0}(t)}{t} &=& p_a b_c(t) - \frac{\alpha_c(t)p_sm_o(t)}{q_c} q_{c,0}(t) +
  \mu_c q_{c,1} - p_b q_{c,0}(t) \label{eq:boinc_ode_1} \\
  \pder{q_{c,j}(t)}{t}&=& \frac{\alpha_c(t)p_sm_o(t)}{q_c} (q_{c,j-1}(t)-q_{c,i}(t)) +
  \mu_c (q_{c,j+1}-q_{c,j}) - p_b q_{c,j}(t)   \label{eq:boinc_ode_2}\\
  \pder{q_{c,J_c}(t)}{t}&=& \frac{\alpha_c(t)p_sm_a(t)}{q_c} q_{c,J_c-1}(t)
  -\mu_c q_{c,J_c} - p_b q_{c,J_c}(t)   \label{eq:boinc_ode_3}\\
  \pder{b_c(t)}{t} &=& - p_v b_c(t) + p_b \sum_{j=0}^{J_c} q_{c,j}(t).
  \label{eq:boinc_ode_4}
\end{eqnarray}
where \eqref{eq:boinc_ode_1} and \eqref{eq:boinc_ode_3} hold for each
cluster $c$ and \eqref{eq:boinc_ode_2} holds for each  cluster $c$ and
for all
$j\le J_c$.  The cost associated to the action function $\alpha$ is:
\begin{equation}
  \int_{0}^T \sum_{c=1}^C\sum_{j=1}^{J_c} jq_{c,j}(t) + \gamma
  \lp\sum_{c=1}^C \frac{\alpha_c(t)p_sm_o(t)}{q_c} (q_{c,J_c}(t)+b_c(t))
  +\sum_{c=1}^C p_b\sum_{j=1}^{J_c} jq_{c,j}(t) \rp dt
  \label{eq:cost}
\end{equation}
The first part of \eqref{eq:cost} represents the cost induced by the number
of jobs in the system. The second part of \eqref{eq:cost} represents the
cost induced by the losses. The parameter $\gamma$  gives  weight
on the cost induced by the losses.

The HJB problem becomes minimizing \eqref{eq:cost} subject to the
variables $u_a,q_{k,i},b_k$ satisfying Equations~\eqref{eq:boinc_ode_0} to
\eqref{eq:boinc_ode_4}. This  system is made of  $(J+2)C$
ODEs.  Solving the HJB equation numerically  in this case can be
challenging
but remains more tractable than solving the original Bellman  equation
over  $J^{N}$ states. The curse of dimensionality is so acute for the
discrete system
that it  cannot be solved  numerically with more than 10
processors \cite{berten}.


\section{Proofs}
\label{sec-proofs}

\subsection{Details of Scaling Constants}
\label{sec-def-cts}
\bearn
I'_0(N,\alpha)& \eqdef &
   I_0(N) + I(N) K e^{(K - L_1)T}
    \lp\frac{K_{\alpha}}{2} +
    2\lp 1+\min(1/I(N),p)\rp\normsup{\alpha}\rp
\\
J(N,T)&\eqdef& 8 T \lc L_1^2 \lb I_2(N)I(N)^2 + I_1(N)^2 \lp T+ I(N) \rp
  \rb + S^2 \lb  2 I_2(N) + I(N) \lp I_0(N) +
  L_2\rp^2\rb\rc
  \\
 B(N,\delta) &\bydef &I(N) \norm{r}_{\infty} +
  K_r  \lp\delta+I_0(N)T\rp
   \frac{e^{L_1T}-1}{L_1} \nonumber
   \\
   &+&
  \frac{3}{2^{1 \over 3}}
    \lb  \frac{K_r}{L_1} \lp e^{L_1T} -1 + \frac{I(N)}{2} \rp\rb^{2 \over 3}
 \norm{r}_{\infty}^{1\over3}J(N,T)^{1\over3}
 \\
B'(N,\delta)&\eqdef&I(N) \norm{r}_{\infty} +
  K_r  \lb\delta+I'_0(N,\alpha)T\rb
   \frac{e^{L_1T}-1}{L_1} \nonumber
   \\
   &+&
  \frac{3}{2^{1 \over 3}}
    \lb  \frac{K_r}{L_1} \lp e^{L_1T} -1 + \frac{I(N)}{2} \rp\rb^{2 \over 3} \norm{r}_{\infty}^{1\over3}J(N,T)^{1\over3}
\eearn

\subsection{Proof of \thref{theo-m-is-markov}}
\label{sec-markov-om}
We begin with a few general statements. Let $\PP$ be the set of probabilities on $\SS$ and $\mu^N: \SS^N \to \PP$ defined by $\mu^N(x)_i=\frac{1}{N}\sum_{n=1}^N \ind{x_n=i}$ for all $i \in \SS$. Also let $\PP^N$ be the image set of $\mu^{N}$, i.e the set of all occupancy measures that are possible when the number of objects is $N$. The following establishes that if two global states have the same occupancy measure, then they differ by a permutation.
\begin{lem}For all $x,x'\in \SS^N$, if $\mu^{N}(x)=\mu^{N}(x')$ there exists some $\sigma \in \mathfrak{S}^N$ such that $x'=\sigma(x)$.
 \label{lem-meme-mu}
 \end{lem}
\begin{IEEEproof}
By induction on $N$. Its is obvious for $N=1$. Assume the lemma holds for $N-1$ and let $x,x'\in \SS^N$, with $\mu^{N}(x)=\mu^N(x')$. There is at least one coordinate, say $i$, such that $x'_i=x_1$, because there is the same number of occurrences of $s=x_1$ in both $x$ and $x'$. Let $y=x_2...x_N$ and $y'=x'_1...x'_{i-1}x'_{i+1}...x'_N$. Then $\mu^{N-1}(y)=\mu^{N-1}(y')$, therefore there exists some $\tau \in \mathfrak{S}^{N-1}$ such that $y'=\tau(y)$. Define $\sigma$ by $\sigma(1)=i$, $\sigma(j)=\tau(j)+\ind{\tau(j)>i}$, for $j\geq2$, so that $x'=\sigma(x)$. Clearly $\sigma$ is a permutation of $\lc 1,...,N\rc$.
\end{IEEEproof}

Let $f: \SS^N\to E$ where $E$ is some arbitrary set. We say that $f$ is invariant under $\mathfrak{S}^N$ if $f\circ \sigma=f$ for all $\sigma \in \mathfrak{S}^N$. The following results states that if a function of the global state is invariant under permutations, it is a function of the occupancy measure.
 \begin{lem}If $f: \SS^N\to E$ is invariant under $\mathfrak{S}$ then there exists $\bar{f}: \PP^N\to E$ such that $\bar{f}\circ \mu^{N}= f$.
 \label{lem-invariance}
 \end{lem}
\begin{IEEEproof}
Define $\bar{f}$ as follows. For every $m\in \PP^N$ pick some arbitrary $x_0\in(\mu^{N})^{-1}(m)$ and let $\bar{f}(m)=f(x_0)$. Now let $x$, perhaps different from $x_0$, such that $\mu^{N}(x)=m$. By \lref{lem-meme-mu}, there exists some $\sigma \in \mathfrak{S}^N$ such that $x=\sigma(x_0)$ therefore $f(x)=f(x_0)=\bar{f}(\mu^{N}(x))$. This is true for every $m \in \PP^N$ thus $f(x)=\bar{f}(\mu^{N}(x))$ for every $x \in \SS^N$.
\end{IEEEproof}

The sequence of actions $a_k$ is given and $N$ is fixed.
We are thus given a time-inhomogeneous Markov chain $X^N$ on $\SS^N$, with transition kernel $G_k$, $k \in\Nats$, given by $G_k(x,y)=\Gamma^N(x,y,a_k)$, such that for any permutation $\sigma\in \mathfrak{S}^N$ and any states $x,y$ we have \be G_k(\sigma(x),\sigma(y))=G_k(x,y)\label{eq-gt}\ee

Let $\FF(k)$ be the $\sigma-$ field generated by $X^N(s)$ for $s\leq k$ and $\GG(k)$ be the $\sigma-$ field generated by $M^N(s)$ for $s\leq k$. Note that because $M^N=\mu^N \circ X^N$, $\GG(k)\subset \FF(k)$.

Pick some arbitrary test function $\varphi:\SS^N \to\Reals$ and fix some time $k\geq 1$; we will now compute $\espc{\varphi(M^N(k))}{\FF(k-1)}$. Because $M^N$ is a function of $X^N$ and $X^N$ is a Markov chain, $\espc{\varphi(M^N(k))}{\FF(k-1)}$ is a function, say $\psi$, of $X^N(k-1)$. We have, for any fixed $\sigma \in \mathfrak{S}^N$:
\bearn
\psi(x) &\eqdef& \sum_{y \in \SS^N}G_k(x,y) \varphi\lp \mu^N(y)\rp
=\sum_{y \in \SS^N}G_k(x,\sigma(y)) \varphi\lp \mu^N(\sigma(y))\rp\\
&=&\sum_{y \in \SS^N}G_k(x,\sigma(y)) \varphi\lp \mu^N(y)\rp\\
\psi(\sigma(x))&=&\sum_{y \in \SS^N}G_k(\sigma(x),\sigma(y)) \varphi\lp \mu^N(y)\rp
= \sum_{y \in \SS^N}G_k(x,y)) \varphi\lp \mu^N(y)\rp
\eearn where the last equality is by \eref{eq-gt}. Thus $\psi(\sigma(x))=\psi(x)$ and by \lref{lem-invariance} there exists some function $\bar{\psi}$ such that $\psi(x)=\bar{\psi}\lp\mu^N(x)\rp$, i.e.
\be
\espc{\varphi(M^N(k))}{\FF(k-1)}=\bar{\psi}\lp M^N(k-1)\rp
\ee In particular, $\espc{\varphi(M^N(k))}{\FF(k-1)}$ is $\GG(k-1)-$measurable. Now
\bearn
 \espc{\varphi(M^N(k))}{\GG(k-1)}
 &=&\espc{\espc{\varphi(M^N(k))}{\FF(k-1)}}{\GG(k-1)}
 \\
 &=&\espc{\bar{\psi}\lp M^N(k-1)\rp}{\GG(k-1)} = \bar{\psi}\lp M^N(k-1)\rp
\eearn
which expresses that $M^N$ is a Markov chain.

\subsection{Proof of \thref{theo-bounds}}
\label{sec-proof-maintheo}
  The proof is inspired by the method in \cite{benaim2003deterministic}.
The main idea of the proof is to write
\begin{eqnarray*}
  \norm{M^N_{\pi}(k)-\phi_{kI(N)}(m_0,A^N_{\pi})}&\leq&
  \norm{M^N_{\pi}(k)-M^N(0)-\sum_{j=0}^{k-1} f^N(j)}\\ &&+
  \norm{M^N(0)+\sum_{j=0}^{k-1} f^N(j)-\phi_{kI(N)}(m_0,A^N_{\pi})}
\end{eqnarray*}
where $f^N(k)\bydef F^N\lp M^N_{\pi}(k),\pi_k(
M^N_{\pi}(k))\rb$ is the drift at time $k$ if the
empirical measure is $M^N_{\pi}(k)$. The first
part is bounded with high probability using a
Martingale argument (Lemma~\ref{lem-p4}) and the
second part is bounded using an  integral formula.

Recall that $ \bar{M}^N_{\pi}(t) \bydef M^N_{\pi}\lp
\floor{\frac{t}{I(N)}}\rp$, i.e.
$\bar{M}^N_{\pi}\lp k I(N)\rp=M^N_{\pi}(k)$ for
$k \in \Nats$ and $\bar{M}^N_{\pi}$ is piecewise
constant and right-continuous. Let $\Delta^N_{\pi}(k)$
be the number of objects that change state
between time slots $k$ and $k+1$. Thus,
\begin{equation}
 \norm{M^N_{\pi}(k+ 1)-M^N_{\pi}(k)}\leq N^{-1}\sqrt{2}\Delta^N_{\pi}(k)
 \label{eq-one-step-lkkjdf}
\end{equation}
and thus
\begin{equation}
  \norm{\hat{M}^N_{\pi}(t)-\bar{M}^N_{\pi}(t)}\leq N^{-1}\sqrt{2}\Delta^N_{\pi}(k)
  \label{eq-one-step-a2}
\end{equation}
 as well, with $k=\floor{\frac{t}{I(N)}}$. Define
 \begin{eqnarray}
   Z^N_{\pi}(k) &=& M^N_{\pi}(k)-M^N(0) - \sum_{j=
     0}^{k-1} F^N\lp M^N_{\pi}(j),\pi_j( M^N_{\pi}(j))\rp
     \label{eq-def-zn}
 \end{eqnarray}
 and let $\hat{Z}^N_{\pi}(t)$ be the continuous, piecewise linear
 interpolation such that $\hat{Z}^N_{\pi}\lp k I(N)\rp=Z^N_{\pi}(k)$
 for $k \in \Nats$. Recall that
 $A^N_{\pi}(t)\bydef\pi_{\lfloor t/I(N) \rfloor}(M^N(\lfloor t/I(N)
 \rfloor))$ -- $A^N_{\pi}(t)$ is the action taken by the controller at
 time $t/I(N)$. It follows from these definitions
 that:
\begin{eqnarray*}
  \hat{M}^N_{\pi}(t) & = &
  M^N_{\pi}(0) +
  \int_0^t \frac{1}{I(N)}F^N\lp \bar{M}^N_{\pi}(s), A^N_{\pi}(s) \rp ds +
  \hat{Z}^N_{\pi}(t)
  \\
  &= &M^N_{\pi}(0) + \int_0^t \frac{1}{I(N)}F^N\lp \hat{M}^N_{\pi}(s),
  A^N_{\pi}(s) \rp ds +
  \hat{Z}^N_{\pi}(t)
  \\
  & &
  + \int_0^t \frac{1}{I(N)}
  \left[
    F^N\lp \bar{M}^N_{\pi}(s), A^N_{\pi}(s)\rp
    -F^N\lp \hat{M}^N_{\pi}(s), A^N_{\pi}(s)\rp
  \right] ds
\end{eqnarray*}

Using the definition of the semi-flow $\phi_t(m_0,A^N_{\pi}) =
m_0 + \int_0^t f(\phi_{s}(m_0,A^N_{\pi}),A^N_{\pi}(s)) ds $, we
get:
\begin{eqnarray*}
 \hat{M}^N_{\pi}(t)-\phi_{t}(m_0,A^N_{\pi})
  &=&
  M^N_{\pi}(0)-m_0
 +\hat{Z}^N_{\pi}(t)
 \\
 & & +  \int_0^t \frac{1}{I(N)}
   \left[
    F^N\lp \hat{M}^N_{\pi}(s), A^N_{\pi}(s) \rp
      -
  F^N\lp \phi_{s}(m_0,A^N_{\pi}),A^N_{\pi}(s)\rp
\right]ds
\\
 & &
 + \int_0^t
  \left[\frac{1}{I(N)}
     F^N\lp \phi_{s}(m_0,A^N_{\pi}), A^N_{\pi}(s)\rp
    -f\lp \phi_{s}(m_0,A^N_{\pi}), A^N_{\pi}(s)\rp
  \right] ds
\\
 & &
 + \int_0^t \frac{1}{I(N)}
  \left[
     F^N\lp \bar{M}^N_{\pi}(s), A^N_{\pi}(s)\rp
    -F^N\lp \hat{M}^N_{\pi}(s), A^N_{\pi}(s)\rp
  \right] ds
\end{eqnarray*}
Applying Assumption~(A2) to the third line, (A3) to the second
and fourth lines, and Equation~\eqref{eq-one-step-a2} to the
fourth line leads to:
\begin{eqnarray*}
  \norm{\hat{M}^N_{\pi}(t)-\phi_t(m_0,A^N_{\pi})}
  &\leq&
  \norm{M^N_{\pi}(0)-m_0}
  +
  \norm{\hat{Z}^N_{\pi}(t)}
  + L_1
  \int_0^t\norm{\hat{M}^N_{\pi}(s)-\phi_s(m_0,A^N_{\pi})}ds
  \\
  & & + I_0(N)t
  +  \frac{\sqrt{2 }L_1 I(N)}{N}
  \sum_{k=0}^{\floor{\frac{t}{I(N)}}}\Delta^N_{\pi}(k)
\end{eqnarray*}
For all $N$, $\pi$, $T$, $b_1>0$ and $b_2>0$, define
\begin{equation}
  \Omega_1  = \lc \omega \in
  \Omega : \sup_{0\leq k \leq \frac{T}{I(N)}} \sum_{j=0}^k \Delta^N_{\pi}(j)
  > b_1 \rc,
  \;
  \Omega_2  =   \lc \omega \in
  \Omega : \sup_{0\leq k \leq
    \frac{T}{I(N)}}\norm{Z^N_{\pi}(k)}>
  b_2\rc
  \label{eq:omega_1}
\end{equation}
Assumption~(A1) implies conditions on the first and second order moment of
$\Delta^N_\pi(k)$. Therefore by \lref{lem-wk}, this shows that for any
$b_1>0$:
\begin{eqnarray}
  \P\lp \Omega_1 \rp &\leq& \frac{T N^2}{b_1^2}\lb I_2(N) +
  \frac{I_1(N)^2}{I(N)^2} \lp T + I(N) \rp\rb \label{eq-pr-o1}
\end{eqnarray}
Moreover, we show in \lref{lem-p4} that:
\begin{eqnarray}
 \P\lp\Omega_2\rp &\leq& 2 S^2 \frac{T}{b^2_2} \lb 2 I_2(N) +
 I(N) \lb \lp I_0(N)+ L_2  \rp
 \rb ^2\rb\label{eq-pr-o2}
\end{eqnarray}
Now fix some $\eps >0$ and let $b_1=\frac{N\eps}{2
\sqrt{2}L_1 I(N)}$, $b_2=\eps/2$. For $\omega \in
\Omega \setminus \lp \Omega_1 \cup \Omega_2\rp$
and for $0\leq t \leq T$: 
\begin{eqnarray*}
  \norm{\hat{M}^N_{\pi}(t)-\phi_t(m_0,A^N_{\pi})}
  &\leq&
  \norm{M^N_{\pi}(0)-m_0} +\eps+ I_0(N)T
  \\
  & & + L_1
  \int_0^t\norm{\hat{M}^N_{\pi}(s)-\phi_s(m_0,A^N_{\pi})}ds
\end{eqnarray*}
By Gr\"onwall's lemma:
 \be
 \norm{\hat{M}^N_{\pi}(t)-\phi_t(m_0,A^N_{\pi})}
 \leq
    \lb
      \norm{M^N_{\pi}(0)-m_0}  + \eps+ I_0(N)T
    \rb
    e^{L_1 t}
    \label{eq-gro}
 \ee
 and this is true for all $\omega \in
\Omega \setminus \lp \Omega_1 \cup \Omega_2\rp$.
We apply the union bound $\P\lp \Omega_1 \cup
\Omega_2\rp\leq
\P\lp\Omega_1\rp+\P\lp\Omega_2\rp$ which, with
\eref{eq-pr-o1} and \eref{eq-pr-o2}, concludes
the proof.

The proof of \thref{theo-bounds} uses the
following lemmas.

\begin{lem}
\label{lem-wk}
Let $\lp W_k\rp_{k\in \Nats}$ be a sequence of
square integrable, non-negative random variables,
adapted to a filtration $\lp \calF_k\rp_{k\in
\Nats}$, such that $W_0=0$ a.s. and for all $k
\in \Nats$:
$\espc{W_{k+1}}{\calF_k}  \leq  \alpha$ and 
$\espc{W_{k+1}^2}{\calF_k}  \leq  \beta$.
 Then for all $n \in \Nats$ and
$b>0$:
 \be
 \P\lp \sup_{0 \leq k \leq n} \lp W_0+...+W_k \rp
 >
 b\rp \leq \frac{n\beta + n(n+1) \alpha^2}{b^2}
 \ee
\end{lem}
\begin{IEEEproof} Let $Y_n= \sum_{k=0}^n W_k$. It follows that
  $\esp{Y_n}\leq \alpha n$ and
    \ben
 \esp{Y_{n+1}^2} \leq \beta + 2n\alpha^2 + \esp{Y_n^2}
 \een from where we derive that
 \be
  \esp{Y_n^2}\leq n\beta + n(n+1) \alpha^2
  \label{eq-dslfkjhdflkjwue}
 \ee
Now, because $W_{n+1}\geq0$: \ben
\espc{Y_{n+1}^2}{\calF_n}\geq \lp\espc{Y_{n+1}}{\calF_n}\rp^2 =\lp Y_n+\espc{W_{n+1}}{\calF_n}\rp^2\geq  Y_n^2
\een thus $Y_n^2$ is a non-negative sub-martingale and by Kolmogorov's inequality:
\ben P\lp\sup_{0\leq k\leq n}Y_k >b\rp=P\lp\sup_{0\leq k\leq n}Y_k^2 >b^2\rp\leq \frac{\esp{Y_n^2}}{b^2}
\een
Together with \eref{eq-dslfkjhdflkjwue} this concludes the proof.

\end{IEEEproof}
%

%
%

\begin{lem}Define $Z^N_{\pi}$ as in
\eref{eq-def-zn}. For all $N \geq 2$, $b >0$,
$T>0$ and all policy $\pi$:
  \begin{equation*}
    \P\lp
    \sup_{0\leq k \leq \floor{T \over I(N)}}  \norm{Z^N_{\pi}(k)} > b\rp
    \leq 2 S^2 \frac{T}{b^2} \lb 2 I_2(N) +
 I(N) \lb \lp I_0(N)+ L_2  \rp
 \rb ^2\rb
  \end{equation*}
  \label{lem-p4}
\end{lem}
\begin{IEEEproof}
The proof is inspired by the methods in
\cite{benaim1999dsa}. For fixed $N$ and $h \in
\Reals^S$, let
 \ben
 L_k = \cro{h, Z^N_{\pi}(k)}
 \een
By the definition of $Z^N$, $L_k$ is a martingale
w.r. to the filtration $\lp\calF_k\rp_{k\in
\Nats}$ generated by $M^N_{\pi}$. Thus
  \bearn
  \espc{\lp L_{k+1}-L_k\rp^2}{\calF_k}& = &
  \espc{\cro{h, M^N_{\pi}(k+1)-M^N_{\pi}(k)}^2}{\calF_k} + \cro{h,F^N\lp
 M^N_{\pi}(k), \pi_k(M^N_{\pi}(k))\rp }^2
  \eearn
  By Assumption (A2):
  \begin{equation*}
    \abs{\cro{h,F^N\lp
        M^N_{\pi}(k), \pi(M^N_{\pi}(k))\rp}} \leq \lp I_0(N)+ L_2  \rp
    I(N) \norm{h}
  \end{equation*}
  Thus, using  \eref{eq-one-step-lkkjdf} and Assumption~(A1):
 \bearn
\espc{\lp L_{k+1}-L_k\rp^2}{\calF_k}& \leq &
 \norm{h}^2 \lb N^{-2}2\espc{\Delta^N_{\pi}(k)^2}{\calF_k}
  +
 \lb \lp I_0(N)+ L_2  \rp
 I(N) \rb ^2\rb
  \\
  & \leq & \norm{h}^2 \lb 2 I(N) I_2(N) +
 \lb \lp I_0(N)+ L_2  \rp
 I(N) \rb ^2\rb
 \eearn

We now apply Kolmogorov's inequality for
martingales and obtain
 \bearn
 \P\lp
   \sup_{0\leq k \leq n} L_k > b
 \rp
 \leq
 \frac{n}{b^2}\norm{h}^2 \lb 2 I(N) I_2(N) +
 \lb \lp I_0(N)+ L_2  \rp
 I(N) \rb ^2\rb
 \eearn

%
%
%
%
%

 Let $\Xi_{h}$ be the set of $\omega \in \Omega$
 such that $\sup_{0\leq k \leq n}  \cro{h, Z^N_{\pi}(k)}   \leq b$ and let
 $\Xi :=\bigcap_{h=\pm \vec{e}_i,
 i=1...S}\Xi_{h}$ where $\vec{e}_i$ is the $i$th vector
 of the canonical basis of $\Reals^S$.
 It follows that, for all $ \omega \in \Xi$ and $0\leq k \leq n$ and
 $i=1\dots S$:
 $\abs{\cro{Z^N_{\pi}(k),\vec{e}_i}}  \leq  b$. This means that for all
 $\omega\in\Xi$: $\norm{Z^N_{\pi}(k)}\leq \sqrt{S} b$.
 By the union bound applied to the complement of $\Xi$,
 we have
 \ben 1- \P(\Xi) \leq 2 S \frac{n}{b^2} \lb I(N) I_2(N) +
 \lb \lp I_0(N)+ L_2  \rp
 I(N) \rb ^2\rb \een
Thus we have shown that, for all $b >0$:
\begin{equation*}
  \P\lp
  \sup_{0\leq k \leq n}  \norm{Z^N_{\pi}(k)}
  >
  \sqrt{S}b\rp \leq 2 S \frac{nI(N)}{b^2} \lb I_2(N) +
  I(N)\lb \lp I_0(N)+ L_2  \rp \rb ^2\rb
\end{equation*}
 which, by changing $b$ to $b/\sqrt{S}$, shows the result.
\end{IEEEproof}

\subsection{Proof of Theorem \ref{th:conv_reward2}}
\label{proof:conv_reward}
 We use the same notation as in the proof of Theorem~\ref{theo-bounds}.
 By definition of $V^N$, $v$ and the time horizons:
 \bearn
 V^{N}_\pi(M^N(0)) - \esp{ v_{A^N_{\pi}}(m_0) }
 & = &
  \esp{ \int_0^{H^N I(N)}
   r(\bar{M}^N_{\pi}(s),  A^N_\pi(s))
    -
   r(m_{A^N_\pi}(s), A^N_\pi(s))
  ds
  } \\
  && -\esp{ \int_{H^N I(N)}^Tr(m_{A^N_\pi}(s), A^N_\pi(s))ds
  }
 \eearn
The latter term is bounded by $I(N) \norm{r}_{\infty}$. Let
$\epsilon>0$ and $\Omega_0=\Omega_1\cup\Omega_2$ where
$\Omega_1,\Omega_2$ are as in the proof of \thref{theo-bounds}.
Thus $\P(\Omega_0)\leq\frac{J(N,T)}{\eps^2}$ and, using the
Lipschitz continuity of $r$ in $m$ (with constant $K_{r}$):
\begin{eqnarray*}
 \abs{V^{N}_\pi(M^N(0)) - \E \lb v_{A^N_{\pi}}(m_0) \rb}
  \leq I(N) \norm{r}_{\infty} +
 \frac{ 2 \norm{r}_{\infty} J(N,T)}{\eps^2}
  +\\
  K_{r}\E \lb1_{\omega \nin  \Omega_0} \int_0^T  \norm{\bar{M}^N_{\pi}(s)-
      m_{A^N_\pi}(s)} ds  \rb
 \end{eqnarray*}
For $\omega \nin \Omega_0$ and $s \in [0,T]$:
$\int_0^T\norm{\bar{M^N_\pi}(s)-\hat{M^N_\pi}(s)}ds\leq
{\epsilon I(N)\over 2 L_1}$ and, by \eref{eq-gro},\\
$\int_0^T\norm{\hat{M}^N_{\pi}(s)-
      m_{A^N_\pi}(s)}ds\leq \lp\norm{M^N(0)-m_0}+I_0(N)T+\epsilon\rp \frac{e^{L_1T}-1}{L_1}$
      thus
\be
   \abs{V^{N}_\pi(M^N(0)) - \E \lb v_{A^N_{\pi}}(m_0) \rb}
  \leq
 B_{\eps}(N,\norm{M^N(0)-m_0})
 \label{eq-inf-be}
\ee where
\begin{equation*}
    B_\epsilon(N,\delta) \bydef I(N) \norm{r}_{\infty}+ K_r  \lp\delta+I_0(N)T+\epsilon\rp
   \frac{e^{L_1T}-1}{L_1}+\frac{K_r I(N)}{2 L_1}  \epsilon +  \frac{2\norm{r}_{\infty}J(N,T)}{\epsilon^2}
  \end{equation*}
This holds for every $\eps >0$, thus \be
   \abs{V^{N}_\pi(M^N(0)) - \E \lb v_{A^N_{\pi}}(m_0) \rb}
  \leq
 B(N,\norm{M^N(0)-m_0})
 \label{eq-inf-be2}
\ee where $B(N,\delta) \bydef
\inf_{\eps>0}B_\epsilon(N,\delta)$.
 By direct calculus, one finds that $\inf_{\eps>0}\lp a \eps +
b/\eps^2\rp=3/2^{2\over3}a^{2\over3}b ^{1\over3}$ for $a>0,
b>0$, which gives the required formula for $B(N, \delta)$.


\subsection{Proof of Theorem \ref{theo:conv_reward1}}
\label{sec:coro1_proof}
Let $\bar{\alpha}^N$ be the right-continuous function constant
on the intervals $[kI(N);(k+1)I(N))$ such that
$\bar{\alpha}^N(s)=\alpha(s)$. $\bar{\alpha}^N$ can be viewed
as a policy independent of $m$. Therefore, by
Theorem~\ref{theo-bounds}, on the set $\Omega
\setminus(\Omega_1 \cup \Omega_2)$, for every $t \in [0;T]$:
 \bearn
  \norm{
 \hat{M}_{\alpha}(t)
 -
 \phi_t(m_0,\alpha)
 }
& \leq&
 \lb
    \norm{M^N(0)-m_0}  + I_0(N) T+ \eps
    \rb
    e^{L_1 T}
    + u(t)
 \eearn
 with
 $u(t)\bydef\abs{\phi_t(m_0,\bar{\alpha}^N)-\phi_t(m_0,\alpha)}$.
 We have
  \begin{eqnarray*}
    u(t) 
    &\leq& \int_{0}^t \abs{f(\phi_s(m_0,\alpha),\alpha(s)) -
      f(\phi_s(m_0,\bar{\alpha}^N),\bar{\alpha}^N(s))}ds \\
    &\leq& \int_{0}^t K \lp\norm{\phi_s(m_0,\alpha) -
      \phi_s(m_0,\bar{\alpha}^N)}+d(\alpha(s),
      \bar{\alpha}^N(s)\rp ds \\
    &\leq& K\int_0^t u(s)ds + K d_1
  \end{eqnarray*}
  where
  $d_1 \bydef\int_0^T\norm{\alpha(t)-\bar{\alpha}^N(t)}dt$.
  Therefore, using Gr\"onwall's inequality, we have
  $ u(t) \leq Kd_1 e^{KT}$.
  By Lemma~\ref{lem:piecewise_alpha}, this shows \eref{eq-q-conv-p2}.
The rest of the proof is as for \thref{th:conv_reward2}.
%
%
%
%
%

\begin{lem} If $\alpha$ is a piecewise Lipschitz continuous
action function on $[0;T]$, of constant $K_\alpha$, and with at
most $p$
  discontinuity points, then
  \label{lem:piecewise_alpha}
  \[ \int_0^Td(\alpha(t),\bar{\alpha}^N(t))dt \leq T I(N)
  \lp\frac{K_\alpha}{2} +  2\lp1+\min(1/I(N),p)\rp\normsup{\alpha}\rp.
  \]
\end{lem}

\begin{IEEEproof}[Proof of lemma \ref{lem:piecewise_alpha}] Let first assume
  that $T=kI(N)$. The left handside
  $d_1=\int_0^T d(\alpha(t),\bar{\alpha}^N(t))dt $ can be decomposed on all
  intervals $[iI(N),(i+1)I(N))$:
  \begin{eqnarray*}
    d_1&=& \sum_{i=0}^{\lfloor T/I(N) \rfloor}  \int_{iI(N)}^{(i+1)I(N)}
    \norm{\alpha(s)-\bar{\alpha}^N(s)} ds 
    \leq  \sum_{i=0}^{\lfloor T/I(N) \rfloor} \int_{iI(N)}^{(i+1)I(N)}
    \norm{\alpha(s)-\alpha(iI(N))} ds
  \end{eqnarray*}
  If $\alpha$ has no discontinuity point on $\lb iI(N),(i+1)I(N)\rp$, then
  \[ \int_{iI(N)}^{(i+1)I(N)}d(\alpha(s),\alpha(iI(N))) ds \le
  \int_{0}^{I(N)}K_\alpha sds \le K_\alpha 2I(N)^2\]
  If $\alpha$ has one or more discontinuity points on $[iI(N),(i+1)I(N))$,
  then
  \[  \int_{iI(N)}^{(i+1)I(N)}d(\alpha(s)\alpha(iI(N))) ds \le
  \int_{iI(N)}^{(i+1)I(N)} 2\normsup{\alpha} ds \le 2\normsup{\alpha}I(N)\]
  There are at most $\min(1/I(N),p)$ intervals $[iI(N),(i+1)I(N)]$ that have
  discontinuity points which shows that
  \[d_1 \leq T I(N)
  (\frac{K_\alpha}{2} + \min(1/I(N),p) 2\normsup{\alpha}).\]

  If $T\ne kI(N)$, then $T=kI(N)+t$ with $0<t<I(N)$. Therefore, there is an
  additional term of $\int_{kI(N)}^{kI(N)+t}d(\alpha(s),
    \bar{\alpha}^N(s)) ds \le 2\normsup{\alpha}I(N)$.
\end{IEEEproof}

\subsection{Proof of \thref{th:exchange_limit}}
\label{sec-proof-main}
 This theorem is a direct consequence of \thref{theo:conv_reward1}
  and Theorem~\ref{th:conv_reward2}. We do the proof for almost sure convergence, the proof for
  convergence in probability is similar. To prove the theorem we prove
 \begin{equation}
    \limsup_{N\to\infty}V^{N}_*(M^N(0)) \le v_*(m_0) \le \liminf_{N\to\infty}
    V^N_*(M^N(0)) \label{eq-mp-2s}
  \end{equation}
  \begin{itemize}
  \item Let $\epsilon>0$ and $\alpha(.)$ be an action
      function such that $v_\alpha(m_0) \geq
      v_*(m_0)-\epsilon$ (such an action is called
      $\epsilon-$optimal). \thref{theo:conv_reward1} shows
      that $\lim_{N\to\infty} V^{N}_\alpha(M^N(0)) =
      v_{\alpha}(m_0) \geq v_*(m_0)-\epsilon$ a.s. This
      shows that $\liminf_{N\to\infty} V^N_*(M^N(0))\geq
      \lim_{N\to\infty} V^{N}_\alpha(M^N(0))\geq
      v_*(m_0)-\epsilon$; this holds for every $\eps>0$
      thus $\liminf_{N\to\infty} V^N_*(M^N(0))\geq
      v_*(m_0)$ a.s., which establishes the second
      inequality in \eref{eq-mp-2s}, on a set of
      probability 1.
  \item Let $B(N,\delta)$ be as in
      Theorem~\ref{th:conv_reward2}, $\epsilon>0$ and
      $\pi^N$ such that $V^{N}_*(M^N(0))\leq
      V^N_{\pi^N}(M^N(0))+ \eps$. By
      Theorem~\ref{th:conv_reward2}, $V^N_{\pi^N}(M^N(0))
      \leq \esp{v_{A^N_{\pi^N}}(m_0)}+B(N,\delta^N) \le
      v_*(m_0)+B(N,\delta^N)$ where
      $\delta^N\bydef\norm{M^N(0)-m_0}$. Thus
      $V^{N}_*(M^N(0))\leq v_*(m_0)+B(N,\delta^N) + \eps$.
      If further $\delta^N\to 0$ a.s. it follows that
      $\limsup_{N\to\infty}V^{N}_*(M^N(0)) \leq v_*(m_0)+
      \eps$ a.s. for every $\eps>0$, thus
      $\limsup_{N\to\infty}V^{N}_*(M^N(0)) \leq v_*(m_0)$
a.s.
      %
%
  \end{itemize}%


\section{Conclusion and Perspectives}
\label{sec-conc}

There are several natural questions arising from this work.
One concerns the convergence of optimal policies.
Optimal policies   $\pi^N_*$  of a  stochastic systems with $N$ objects
may not be unique, they  may also exhibit thresholds and therefore  be discontinuous. This implies that  $M^N_{\pi^N_*}$ and $V^N_{\pi^N_*}$  will  not  converge in general. In some particular cases, such as the best response
dynamics studied in \cite{gorodeisky2009deterministic}, limit
theorems can nevertheless  be obtained, at the cost of a much greater complexity. In full generality however, this problem  is still open and definitely deserves further investigations. 

The second question concerns the time horizon.
In this paper we have focused  on the finite horizon case.
Actually,  most  results  and in particular theorems \ref{th:exchange_limit} and \ref{theo:conv_reward1},
remain valid with an infinite horizon with discount.
The main argument that makes everything work in the discounted case is the following.  When the rewards $r(s,a)$ are  bounded, for a given discount $\beta < 1$ and a given $\varepsilon >0$, it is possible to find a finite time horizon $T$ such that the expected discounted value of a policy $\pi$ can be decomposed into the value over time $T$ plus a term less than $\varepsilon$:

$$\E \sum_{t>0} \beta^t r(M^N(t),\pi(M^N(t)) \leq  \E \sum_{t=0}^T \beta^t r(M^N(t),\pi(M^N(t)) + \varepsilon.$$
Therefore, the main result of this paper, which states that a policy $\pi$ that is optimal in the mean field limit is near-optimal  for the finite system with $N$ objects, also holds in the infinite horizon discounted case.

As for the infinite horizon without discount or average reward cases, convergence of the value when $N$ goes to infinity is not guaranteed in general. Finding natural assumptions under which convergence holds  is also one of our goals  for the future.

\bibliographystyle{plain}
\bibliography{bibFile}

\end{document}